%% file: main.tex
\documentclass[10pt,twocolumn,letterpaper,table,dvipsnames]{article}

\usepackage[many]{tcolorbox}

\newcommand{\up}[1]{\textcolor{red}{(+#1)}}

\usepackage{arydshln}
\usepackage[hypcap=false]{caption}
\usepackage[pagenumbers]{iccv} 

\input{preamble}

\definecolor{iccvblue}{rgb}{0.21,0.49,0.74}
\usepackage[pagebackref,breaklinks,colorlinks,allcolors=iccvblue]{hyperref}

\newcommand{\ours}{FinMMR\xspace}
\renewcommand{\github}{\raisebox{-1.5pt}{\includegraphics[height=1em]{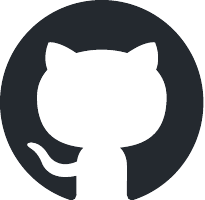}}}
\renewcommand{\huggingface}{\raisebox{-1.5pt}{\includegraphics[height=1em]{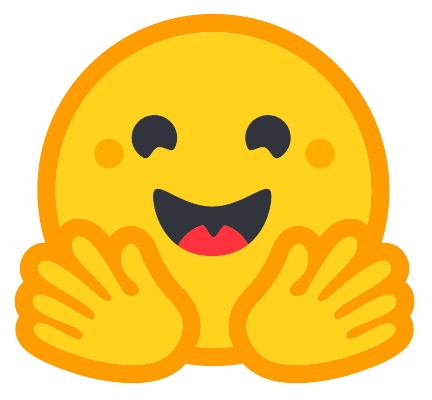}}}
\newcommand{\logo}{%
    \raisebox{-0.7ex}{
    \includegraphics[width=2em,height=2em,keepaspectratio]{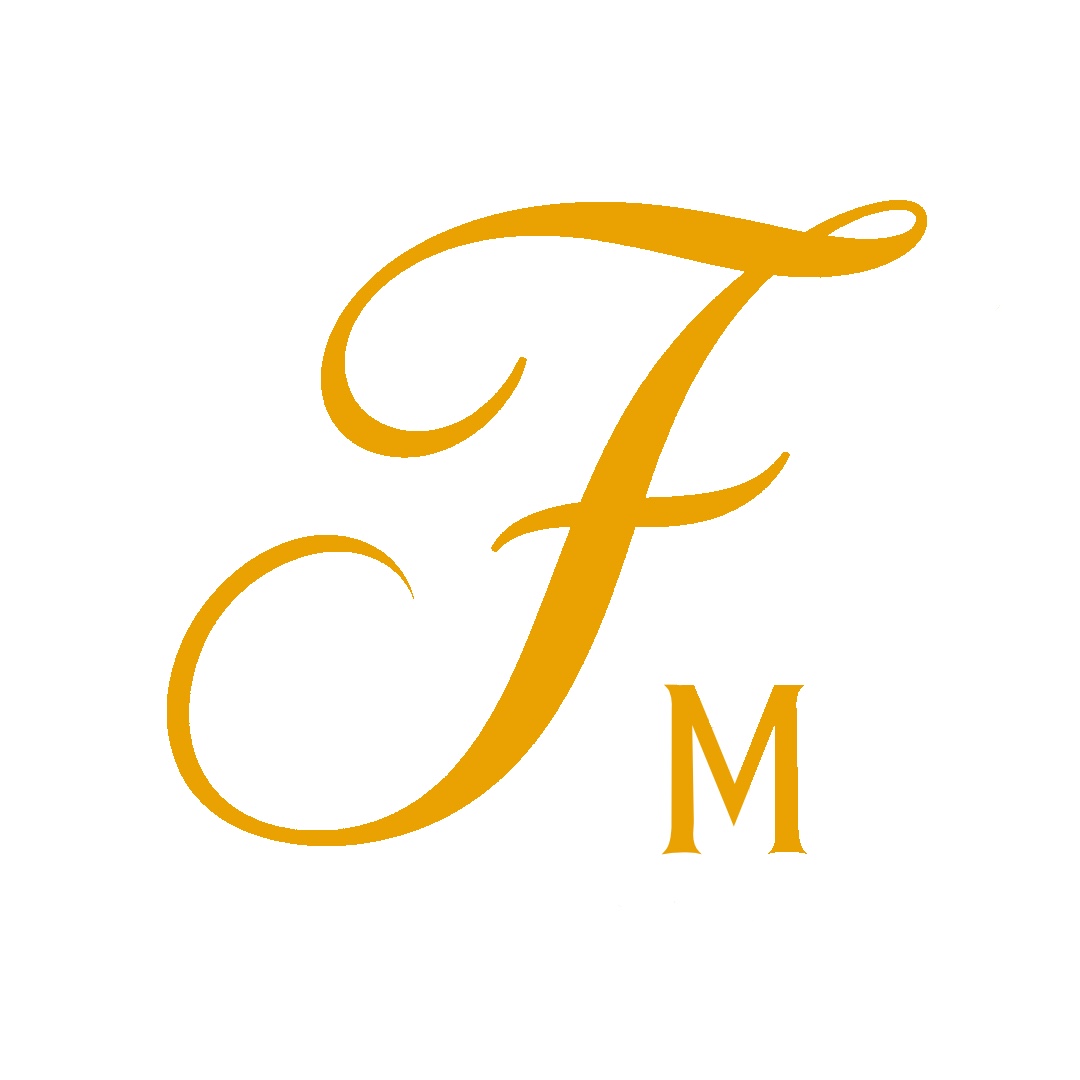}%
    }%
}

\def\paperID{15969} 
\def\confName{ICCV}
\def\confYear{2025}

\title{\logo\ours: Make Financial Numerical Reasoning \\More Multimodal, Comprehensive, and Challenging}

\author{
Zichen Tang \quad 
Haihong E\thanks{Corresponding author.} \quad 
Jiacheng Liu \quad 
Zhongjun Yang \quad
Rongjin Li \quad
Zihua Rong \\
Haoyang He \quad 
Zhuodi Hao \quad
Xinyang Hu \quad 
Kun Ji \quad
Ziyan Ma \quad
Mengyuan Ji \quad 
Jun Zhang \\
Chenghao Ma \quad
Qianhe Zheng \quad
Yang Liu \quad
Yiling Huang \quad
Xinyi Hu \quad
Qing Huang \quad
Zijian Xie \\
Shiyao Peng \vspace{4pt}\\
Beijing University of Posts and Telecommunications\\\newline \vspace{10pt}
}

\begin{document}
\maketitle

\input{sec/0_abstract}    
\input{sec/1_introduction}
\input{sec/5_relatedwork}
\input{sec/2_benchmark}

\input{sec/3_evaluation}
\input{sec/4_experiments}

\input{sec/6_conclusion}
\input{sec/acknowledgements}
{
    \small
    \bibliographystyle{ieeenat_fullname}

\input{main.bbl}
}

\begin{onecolumn}

\appendix
\clearpage
\input{error/main}
\input{augmentation/main}
\input{prompt_experiment/main}

\input{error/error_type}
\input{error/visual_perception_error}
\input{error/knowledge_reasoning_error1}
\input{error/knowledge_reasoning_error2}
\input{error/numerical_calculation_errors}

\input{augmentation/pic_insert}

\input{prompt_experiment/table_model}
\input{prompt_experiment/table_validation_result}
\input{prompt_experiment/prompt_cot}
\input{prompt_experiment/prompt_pot}
\input{prompt_experiment/prompt_rag}
\input{prompt_experiment/prompt_judge_useful_function}
\input{prompt_experiment/prompt_judge_correlation_of_image}

\end{onecolumn}
\end{document}

%% file: preamble.tex
\usepackage{multirow}   
\usepackage{booktabs}   
\usepackage{booktabs}   
\usepackage{array}      

\usepackage{pifont} 
\definecolor{darkgreen}{rgb}{0,0.5,0}     
\usepackage{makecell}  

%% file: sec/0_abstract.tex
\begin{abstract}
We present \textbf{\ours}, a novel bilingual multimodal benchmark tailored to evaluate the reasoning capabilities of multimodal large language models (MLLMs) in financial numerical reasoning tasks. Compared to existing benchmarks, our work introduces three significant advancements. (1) \textbf{Multimodality}: We meticulously transform existing financial reasoning benchmarks, and construct novel questions from the latest Chinese financial research reports. \ours comprises 4.3K questions and 8.7K images spanning 14 categories, including tables, bar charts, and ownership structure charts. (2) \textbf{Comprehensiveness}: \ours encompasses 14 financial subdomains, including corporate finance, banking, and industry analysis, significantly exceeding existing benchmarks in financial domain knowledge breadth. (3) \textbf{Challenge}: Models are required to perform multi-step precise numerical reasoning by integrating financial knowledge with the understanding of complex financial images and text. The best-performing MLLM achieves only 53.0\% accuracy on Hard problems. We believe that \ours will drive advancements in enhancing the reasoning capabilities of MLLMs in real-world scenarios.

\end{abstract}


%% file: sec/1_introduction.tex
\begin{figure*}[!t]
    \centering
\includegraphics[width=\linewidth]{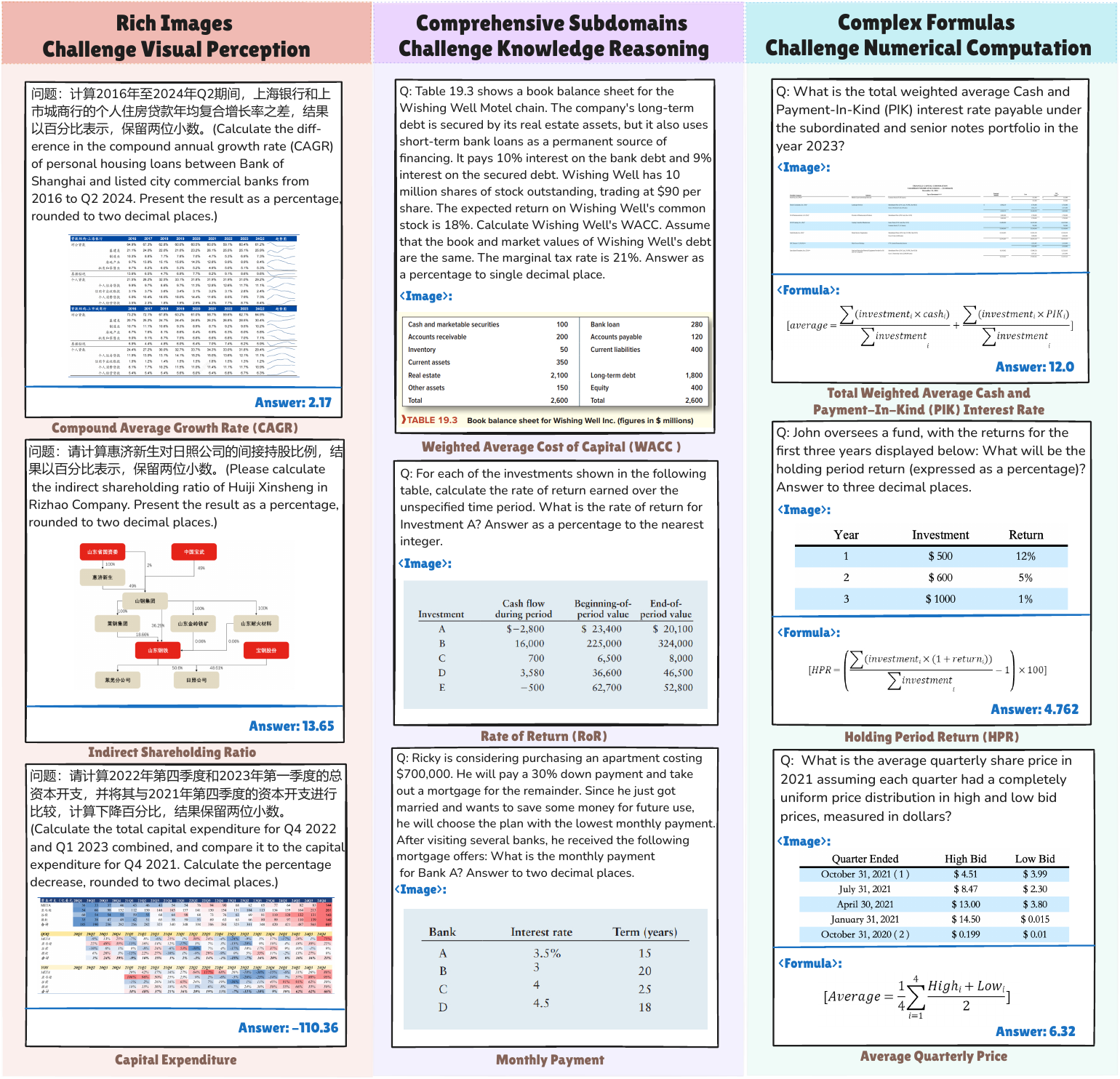}
    \caption{Sampled \ours examples with two language (\ie{, English and Chinese}), rich images and different knowledge. The questions and images need expert-level visual perception, knowledge reasoning and numerical computation.}
    \label{fig:finmmr_examples}
\end{figure*}

\section{Introduction}
Recently, large reasoning models (LRMs)~\cite{xu2025towards,openai2024o1mini,openai2025o3mini,guo2025deepseekr1,qwq-32b-preview,team2025kimi}, show powerful reasoning capabilties over multi-step reasoning tasks, with train-time scaling and test-time scaling~\cite{kaplan2020scaling,openai2024learning}. These reasoning models are proficient in code~\cite{jain2024livecodebench, chen2021humaneval}, math~\cite{mao-etal-2024-champ, lightman2023let}, and science~\cite{wang2024scibench}. Multimodal large language models (MLLMs) ~\cite{openai2024o1,gemini-2.0-flash-thinking,claude-3-7-sonnet} also exhibit greater performance on multimodal reasoning~\cite{lu2023mathvista, yue2023mmmu}.

Despite significant advancements, there remains a notable gap in understanding the practical applicability of MLLMs in numerical reasoning within real-world scenarios, particularly in high-stakes fields such as finance and healthcare. As shown in ~\cref{fig:overview}, financial analysts in their daily work are required to read visually enriched financial documents, extract key financial indicators from tables, images, and texts, and perform multi-step precise numerical calculations to support professional decision-making. Similarly, to achieve expert artificial general intelligence (AGI)~\cite{bubeck2023sparksartificialgeneralintelligence,ge2023openagi,morris2024position,mialon2024gaia,yue2023mmmu}, MLLMs are expected to comprehend complex domain-specific images akin to human experts, and apply domain knowledge to perform accurate numerical reasoning. This raises the question: \textbf{Can current MLLMs seamlessly integrate vision and text to perform domain-specific complex reasoning, matching the proficiency of LRMs in pure text-based tasks?}

Specifically, we choose the financial domain to evaluate the complex reasoning capabilities of MLLMs, where precision and transparent reasoning are paramount~\cite{krumdick-etal-2024-bizbench}. Existing numerical reasoning benchmarks for finance are limited in their text-based reasoning, coverage of specific financial knowledge, and complexity of reasoning~\cite{chen-etal-2021-finqa, chen-etal-2022-convfinqa, zhu-etal-2021-tat, zhao-etal-2024-knowledgefmath, krumdick-etal-2024-bizbench}. FAMMA~\cite{xue2024fammabenchmarkfinancialdomain} is mainly modelled after textbooks and CFA exam questions, MathVista~\cite{lu2023mathvista} does not involve the application of financial knowledge, MMMU~\cite{yue2023mmmu} and MMMU-Pro~\cite{yue-etal-2025-mmmu}
 are restricted to multiple-choice questions, diverging from authentic financial problem-solving. The lack of high-quality, knowledge-intensive multimodal financial numerical reasoning benchmarks makes it challenging to objectively evaluate the actual reasoning capabilities of MLLMs and analyze their shortcomings.

Therefore, we propose \textbf{\ours}, a bilingual multimodal numerical reasoning benchmark designed to evaluate the reasoning capabilities of MLLMs in the finance domain. The benchmark comprises 4.3K problems, covering 14 financial subdomains (\eg{, corporate finance and industry analysis}), with 8.7K images derived from 14 categories (\eg{, tables and ownership structure charts}). Each problem includes rich images, an unambiguous question, a Python-formatted solution, and a precise answer.

\textbf{For multimodality}, without representing financial tables as structured text, \ours represent all tables, charts, and diagrams as images. \textbf{For comprehensiveness}, \ours covers 14 financial subdomains and two languages (\ie{, English and Chinese}), demanding domain knowledge such as option pricing and portfolio management. \textbf{For challenge}, \ours focus on multi-step numerical reasoning, requiring models to provide exact numerical answers under strict evaluation criteria (emphasizing units, percentages, and decimal places). Furthermore, we \textbf{mix each Chinese questions with two distractor images} that are contextually adjacent to the ground images, approaching real-world multimodal reasoning scenarios.

We evaluate 15 state-of-the-art (SOTA) MLLMs~\cite{openai2024o1,openai2024o1mini,openai2025o3mini,guo2025deepseekr1,qwq-32b-preview,gemini-2.0-flash-thinking,gemini-2.0-pro,gemini-2.0-flash}, utilizing Chain-of-Thought (CoT)~\cite{wei2022chain} and Program-of-Thought (PoT)~\cite{chen2023program}. The experimental results on FinMMR reveals three key findings:
\begin{itemize} [leftmargin=*]
    \itemsep 0em 
    \item \textbf{MLLMs Face Significant Challenges in Domain-Specific Multimodal Numerical Reasoning}: 
    All evaluated models underperform on \ours. On the \emph{Hard} \emph{test} set, the best-performing model, Claude 3.7 Sonnet with 64K extended thinking, achieves only 53.00\% accuracy, while OpenAI o1 achieves merely 48.40\%. Through error analysis, we identify that visual perception, knowledge reasoning, and numerical computation collectively pose challenges to MLLMs. Current MLLMs still struggle with complex multimodal reasoning tasks in specialized domains, compared to text-based reasoning.

    \item \textbf{Better Synergy Between Visual Perception and Complex Reasoning Is Needed:} 
    Distracting images lead to a greater than 10\% drop in accuracy for Qwen2.5-VL-72B compared to ground images alone, indicating that irrelevant visual information severely impacts multimodal reasoning. By decoupling visual filtering and reasoning, Qwen2.5-VL-72B improved from 64.73\% to 71.56\%. Combining MLLMs with LRMs, by efficiently parsing visual information into structured text and leveraging the LRMs' text-based reasoning capabilities, can also yield better performance. The combination of GPT-4o and DeepSeek-R1 achieves 86.72\% accuracy on 1,160 tabular QA instances, outperforming Claude 3.7 Sonnet (83.53\%).

    \item \textbf{Refined Structured Domain Knowledge Enhances MLLMs' Complex Reasoning:} 
    Leading MLLMs lack sufficient experience in applying rich domain knowledge when solving complex reasoning tasks. By annotating structured financial functions and leveraging the model's ability to generate retrieval questions and make judgments, knowledge augmentation can significantly improve MLLMs' performance. MLLMs achieve absolute improvements ranging from 2.76 to 4.31 percentage points, allowing weaker models to approach SOTA performance, while SOTA models exhibit further gains.
\end{itemize}

These findings highlight the bottlenecks of MLLMs in complex multimodal reasoning tasks in expert domains closer to real-world scenarios. They emphasize the need for continuous improvements in three key areas: more intricate visual perception, more specialized knowledge reasoning, and more accurate numerical computation. Alternatively, leveraging tools or model combinations can help achieve a balance between performance and cost, enabling MLLMs to perform expert-level reasoning tasks like human experts.

%% file: sec/5_relatedwork.tex
\section{Related Work}

\subsection{LRMs and MLLMs}
By integrating train-time scaling and test-time scaling~\cite{kaplan2020scaling, openai2024learning}, LRMs have demonstrated remarkable reasoning capabilities~\cite{xu2025towards}. 
However, most LRMs are limited to handling text-based problems. The growing demand for solving real-world tasks has spurred the development of MLLMs~\cite{claude-3-7-sonnet, qvq-72b-preview, gemini-2.0-flash-thinking} and benchmarks designed to evaluate the perception and reasoning abilities of MLLMs ~\cite{hao2025mllmsreasonmultimodalityemma,yue-etal-2025-mmmu,jiang2025mmecotbenchmarkingchainofthoughtlarge,fu2024mmecomprehensiveevaluationbenchmark,liu2024mmbenchmultimodalmodelallaround,yu2024mmvetevaluatinglargemultimodal,liu-etal-2024-mmc,yue2023mmmu,Li_2024_CVPR,chen2024rightwayevaluatinglarge}. For instance, MME-CoT~\cite{jiang2025mmecotbenchmarkingchainofthoughtlarge} evaluates models' space-time understanding, while EMMA~\cite{hao2025mllmsreasonmultimodalityemma} focuses STEM subjects. Following this trend, domain-specific benchmarks which require deep domain expertise have emerged, such as MathVista~\cite{lu2023mathvista} for mathematics and GMAI-MMBench~\cite{chen2024gmaimmbenchcomprehensivemultimodalevaluation} for medicine. Yet, financial reasoning remains an unexplored area in the current landscape of MLLM benchmarks.

\subsection{Financial Benchmarks}
The financial domain presents a distinct and more formidable set of challenges for model evaluation, which arise from its inherent complexity, information density, and dependence on expertise. The majority of existing text-only financial numerical reasoning benchmarks ~\cite{chen-etal-2021-finqa,zhu-etal-2021-tat,krumdick-etal-2024-bizbench,chen-etal-2022-convfinqa,zhao-etal-2024-knowledgefmath,zhao-etal-2024-docmath} are constrained by limitations such as sub-optimal annotation quality, narrow domain knowledge coverage, and overly simplistic reasoning tasks. Although FinanceReasoning~\cite{tang-etal-2025-financereasoning} offers complex tasks with rich knowledge and high-quality annotations, its text-only modality limits multimodal reasoning evaluation.

Recent multimodal financial benchmarks have sought to bridge this gap but still possess limitations. FAMMA~\cite{xue2024fammabenchmarkfinancialdomain} being sourced from textbooks and examinations does not mirror the real-world tasks. FinMME~\cite{luo-etal-2025-finmme} uses a multiple-choice format, which may overestimate model reasoning due to guesswork. MME-Finance~\cite{gan2024mmefinancemultimodalfinancebenchmark} is constrained by coarse annotations and an isolated assessment of domain knowledge, limiting holistic evaluation of real-world financial reasoning. 

%% file: sec/2_benchmark.tex
\section{\ours Benchmark}

\input{tables/table_key_statistics}
\subsection{Overview of \ours}
We introduce \ours, a bilingual (English and Chinese) multimodal benchmark for evaluating the reasoning capabilities of MLLMs in financial numerical reasoning tasks. \ours consists of 4,300 questions covering 14 financial subdomains (\eg{, corporate finance, industry analysis, financial markets, asset management}). The key statistics are summarized in \cref{tab:dataset_statistics}, and the composition of sub-domains and images is illustrated in \cref{fig:overview}. As illustrated in \cref{fig:finmmr_examples}, \ours introduces three key challenges:



\begin{itemize} [leftmargin=*]
    \itemsep 0em 
    \item \textbf{Rich Images Challenge Visual Perception}: \ours comprises 8.7K images from 14 categories (\eg{, bar charts, line charts, ownership structure charts}). MLLMs must identify relevant images among distractors and extract critical information from the correct images.
    \item \textbf{Comprehensive Subdomains Challenge Knowledge Reasoning:} MLLMs need to flexibly apply diverse domain-specific financial knowledge from 14 subdomains to solve multi-step reasoning tasks.
    \item \textbf{Complex Formulas Challenge Numerical Computation:} All questions require precise numerical answers, eliminating the potential bias from lucky guesses that could occur in multiple-choice formats.
\end{itemize}

\input{tables/table_dataset_comparison}

\subsection{Data Curation Process}
We first curated a subset of questions from public text-based financial reasoning benchmarks and systematically transformed them into multimodal problems using a unified standard. Subsequently, we constructed a novel multimodal Chinese Research Report Question Answering (CRRQA) benchmark from scratch, merging two data sources into \ours. Each question is accompanied by an executable Python solution, yields a numerical answer and demonstrates a clear reasoning chain. 

\noindent \textbf{Update to Public Benchmarks} We re-annotate 124 and 163 financial questions from MMMU~\cite{yue2023mmmu} and MMMU-Pro~\cite{yue-etal-2025-mmmu}, respectively. Following rigorous verification, these questions were directly incorporated into our dataset. Furthermore, we extracted 77, 288, and 795 questions from FinanceMath~\cite{zhao-etal-2024-knowledgefmath}, CodeTAT-QA~\cite{krumdick-etal-2024-bizbench}, and CodeFinQA~\cite{krumdick-etal-2024-bizbench}, respectively. From DocMath-Eval~\cite{zhao-etal-2024-docmath}, we further obtained 703 questions from its four subsets. For each question, we rendered any tabular data as images while removing the corresponding table information from the text, ensuring that MLLMs cannot rely solely on textual content.

\noindent \textbf{Building a Novel Dataset from Scratch} We collect 90 research reports, all of which are obtained through authorized access and cover diverse topics such as industry research, macroeconomic analysis, and strategy research. We use 360LayoutAnalysis~\cite{360LayoutAnalysis} to extract informative images and discard those lacking explicit numerical data, reducing ambiguity. For each retained image, we prompt Qwen-VL-Max~\cite{QwenVL} to formulate questions requiring multiple reasoning steps or complex calculations. Each question is accompanied by an executable Python solution and a definitive numerical answer. 

Furthermore, we introduce distractor images sourced from the same reports adjacent to ground images to challenge MLLMs in extracting relevant numerical information from structured, densely packed visuals.



\noindent \textbf{Data Quality Assurance} This process ensured that every question was clearly written, featured a detailed reasoning solution, and included an accurate final answer. The annotators included 16 graduate students in finance and two experts holding CFA certifications. With the aid of LLMs, this meticulous verification process spanned three months, culminating in a dataset that meets high standards of clarity and correctness.

\noindent \textbf{Dataset Splitting and Release} To classify the problems by difficulty, we employ a heuristic algorithm that takes into account the number of operators (\(o\)), code lines (\(l\)) and parentheses pairs (\(p\)) in the Python solution. Specifically, the difficulty of reasoning \(rc\) of a problem is defined as:
\begin{equation}
    rc = \ln(\max(o,1)) + \ln(\max(l + p,1))
    \label{eq:rc_formula}
\end{equation}

\ours is classified as \emph{Easy} (1,300 examples), \emph{Medium} (1,500 examples) and \emph{Hard} (1,500 examples) based on this formula, with each level randomly split into \emph{test} and \emph{validation} sets. All questions are publicly available, while only the 300 validation answers per level are released, while test answers remain private to prevent data leakage \cite{deng-etal-2024-investigating,shi2024detecting,sainz-etal-2023-nlp}. We maintain an online evaluation platform that enables researchers to assess their models.



\subsection{Comparisons with Existing Benchmarks}
As illustrated in \cref{fig:finance_comparison}, previous work has studied multi-discipline multimodal reasoning~\cite{yue2023mmmu, yue-etal-2025-mmmu}, general mathmatical reasoning~\cite{lu2023mathvista} or text-based financial QA \cite{krumdick-etal-2024-bizbench, zhao-etal-2024-knowledgefmath,zhao-etal-2024-docmath}. \ours focus on multimodal financial numerical reasoning, curating 4,300 questions requiring a deep understanding of domain-specific images (\eg{, earnings reports, candlestick charts}). To mimic real-world scenario, we deliberately incorporate 3,938 distractors into 2,150 questions to rigorously evaluate MLLMs' visual perception capability. Compared to existing financial benchmarks, they suffer from narrow domain coverage~\cite{chen-etal-2021-finqa,zhu-etal-2021-tat}. \ours encompasses 14 financial subdomains and 14 image categories, comprehensively evaluating MLLMs' domain-speific reasoning capabilities.

%% file: tables/table_key_statistics.tex
\begin{table}
\centering
\begin{tabular}{lr}
\toprule
\textbf{Property} & \textbf{Value} \\
\midrule
\# Total Questions & 4,300 \\
\# Easy/Medium/Hard & 1,300/1,500/1,500 \\
\# Validation/Test & 900/3,400 \\
\# Chinese/English & 2,150/2,150 \\
\midrule
\# Operators (Easy/Medium/Hard) & 1.75/2.97/\textbf{5.34} \\
\# Lines of Code (Easy/Medium/Hard) & 4.14/5.06/\textbf{7.34} \\
\# Parentheses (Easy/Medium/Hard) & 0.88/3.11/\textbf{4.25} \\
\# Difficulty (Easy/Medium/Hard) & 1.93/2.96/\textbf{3.79} \\

\bottomrule
\end{tabular}
\caption{Key statistics of \ours (Avg values of three subsets).}
\label{tab:dataset_statistics}
\end{table}

%% file: tables/table_dataset_comparison.tex
\begin{figure*}[thb]
\raggedright 

\begin{minipage}{0.40\linewidth}
\includegraphics[width=0.92\textwidth]{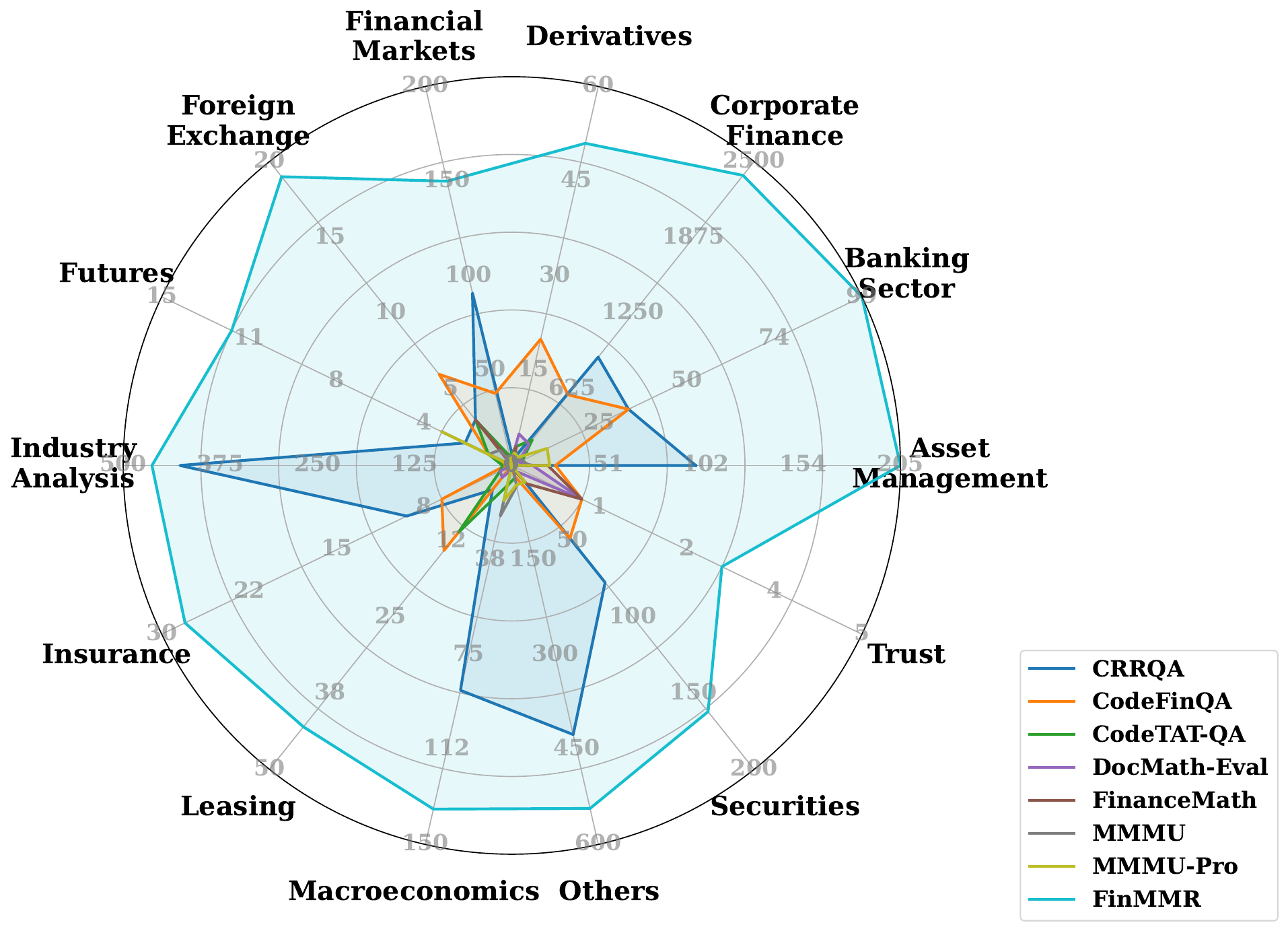}
\end{minipage}
\begin{minipage}{0.5\linewidth}
\captionsetup{type=table} 
{\small
\begin{tabular}{llclll}
\toprule
\multirow{2}{*}{\textbf{Benchmark}} & \multirow{2}{*}{\textbf{Size (Fin)}} & \multirow{2}{*}{\textbf{\makecell[c]{\raisebox{-0.2em}{Domain}\\ \raisebox{-0.2em}{Coverage}}}} & \multirow{2}{*}{\textbf{Modalities}} & \multirow{2}{*}{\textbf{\makecell[c]{\raisebox{-0.2em}{Question}\\ \raisebox{-0.2em}{Type}}}} \\
\\
\midrule
MMMU~\cite{yue2023mmmu} & 11,550 (1,603) & 10 & T+I & MC \\
MMMU-Pro~\cite{yue-etal-2025-mmmu} & 1,730 (286) & 10 & T+I, P.I. & MC \\
FinanceMath~\cite{zhao-etal-2024-knowledgefmath} & 1,200 (1,200) & 10 & T  & NUM \\
CodeTAT-QA~\cite{krumdick-etal-2024-bizbench} & 3,144 (3,144) & 6 & T & NUM \\
CodeFinQA~\cite{krumdick-etal-2024-bizbench} & 5,463 (5,463) & 13  & T & NUM \\
DocMath-Eval~\cite{zhao-etal-2024-docmath} & 4,000 (4,000) & 12  & T & NUM \\
FAMMA~\cite{xue2024fammabenchmarkfinancialdomain} & 1,758 (1,758) & 8 & T+I & MC, NUM\\
CRRQA (ours) & 2,150 (2,150) & 13 & T+I, P.I. & NUM \\
FinMMR (ours) & 4,300 (4,300) & 14 & T+I, P.I. & NUM \\
\bottomrule
\end{tabular}
}
\end{minipage}
\hfill
\caption{The comparison between finance-related benchmarks. These benchmarks vary in size, domain coverage, modalities, and question type, with some focusing on text-only data while others include images. Each axis has scale labels with varying ranges to measure the number of questions from each benchmark across different subdomains. In the Modalities, T means text input, I means Images input, P.I. means pure images input. In the Question Type, MC means multi-choice answer, NUM means numerical answer.}
\label{fig:finance_comparison}
\end{figure*}

%% file: sec/3_evaluation.tex
\section{Evaluation System}
To facilitate the evaluation of complex reasoning on FinMMR, we established a dedicated evaluation system. All MLLMs evaluated were accessed through official APIs. 


\subsection{Multimodal Large Language Models}
We systematically evaluate the multimodal reasoning capabilities of 15 recent MLLMs under the zero-shot setting. The evaluated MLLMs are: OpenAI o1~\cite{openai2024o1}, GPT-4o~\cite{openai2024gpt4o}, Claude 3.7 Sonnet (including thinking mode)~\cite{claude-3-7-sonnet}, Gemini 2.0 Flash Thinking~\cite{gemini-2.0-flash-thinking}, Gemini 2.0 Pro~\cite{gemini-2.0-pro}, Gemini 2.0 Flash~\cite{gemini-2.0-flash}, InternVL2.5-78B~\cite{internvl2.5}, Grok 2 Vision~\cite{grok-2-vision-1212}, Pixtral Large~\cite{pixtral-large}, Qwen2.5-VL-72B~\cite{bai2025qwen25vltechnicalreport}, QVQ-72B-Preview~\cite{qvq-72b-preview}, Qwen-Omni-Turbo~\cite{qwen-omni-turbo}, Llama 4 Maverick~\cite{llama4maverick}, Gemma 3 27B~\cite{Gemma3}, and Mistral Small 3.1~\cite{mistralsmall3.1}.
 

\subsection{Prompting Methods}
Our evaluation adopts CoT~\cite{wei2022chain}, PoT~\cite{chen2023program} and IO (without any external prompts) prompting methods. Due to budget constraints, we report OpenAI o1 performance on the \emph{Hard} subset only. Detailed prompts can be found in the Appendix.

\subsection{Answer Extraction and Evaluation}
Following \citet{zhao-etal-2024-knowledgefmath}, we extract answers based on the prompting methods. For CoT and IO outputs, we employ GPT-4o-mini to extract numerical answers. For PoT, we run the generated program for numerical results. Finally, we conduct a strict accuracy evaluation, comparing numerical results with ground truth and deeming the results accurate within a stringent error tolerance of 0.2\%.

%% file: sec/4_experiments.tex
\input{tables/table_main_result}

\section{Experiments}
We answer the following research questions (RQs):
\textbf{RQ1}: Are MLLMs multimodal reasoners with extended thinking and PoT prompts?
\textbf{RQ2}: What are the primary challenges facing MLLMs?
\textbf{RQ3}: How can the visual perception difficulties of MLLMs be mitigated?
\textbf{RQ4}: How can the knowledge reasoning capabilities of MLLMs be enhanced?
\textbf{RQ5}: How can the numerical computation abilities of MLLMs be compensated for?

\subsection{Main Results (RQ1)}
The performance of the MLLMs evaluated using different prompting methods on FinMMR is shown in \cref{tab:main_result}.

\noindent \textbf{Challenges of MLLMs in Domain-Specific Complex Numerical Reasoning} As the difficulty increases, the average accuracy shows a continuous and significant decline. In CoT setting, the average accuracy rates on the \emph{Easy}, \emph{Medium}, and \emph{Hard} subsets are 73.33\%, 54.05\%, and 38.14\%, respectively. The currently best-performing model (\ie{, Claude 3.7 Sonnet with 64K extended thinking}) consistently demonstrates superior performance across all difficulty subsets using CoT prompting method. \textbf{However, its accuracy on the \emph{Hard} subset remains below the 60\% passing threshold in both prompting methods.} On the overall \emph{test} set, Claude 3.7 Sonnet achieves only 64.02\% accuracy. These results highlight the challenging nature and rigorous standards of FinMMR, effectively assessing the limits of MLLMs' reasoning capabilities and the disparities among models compared to previous benchmarks.



\noindent \textbf{Does extended thinking help?} \textbf{Reasoning-enhanced models show consistent improvements, compared with non-reasoning-enhanced MLLMs.} Claude 3.7 Sonnet with 64K extended thinking achieves 2.20 percentage points higher accuracy than the model without extended thinking (53.00\% vs. 50.80\%) on the \emph{Hard} subset. This enhancement comes at the cost of using nearly 12 times more tokens for intricate thinking (4.06M vs. 0.34M). This trend also persists in the Gemini 2.0 Flash series. 

We observe that QVQ-72B-Preview lose basic code generation capabilities due to the reinforcement learning of text-based long thought, which is likely attributed to biases in training strategies and training data. On the \emph{Hard} subset, this model achieves a code execution success rate of only 10.90\%, resulting in an accuracy of merely 6.20\% in PoT setting, significantly lower than the 40.30\% accuracy achieved with CoT. This finding highlights the importance of maintaining foundational capabilities, such as programming, while enhancing the reasoning abilities of MLLMs, to avoid rendering them ineffective in other basic tasks.


\input{tables/table_vision_filter}
\noindent \textbf{Does PoT help?} Experimental results strongly validate the superiority of PoT over CoT in numerical reasoning tasks, especially on the \emph{Hard} subset. PoT (not including QVQ-72B-Preview) achieves a mean accuracy of 37.64\% versus 36.20\% for CoT, representing an improvement of 1.44 percentage points. Furthermore, PoT encourages MLLMs to leverage structured code generation to reduce token consumption during reasoning. Under similar or lower token usage, PoT achieves similar or greater accuracy. For instance, GPT-4o achieves a 2.40 percentage points improvement in accuracy over CoT while consuming significantly fewer tokens in PoT setting. Similarly, Qwen2.5-VL-72B demonstrates the most pronounced efficiency gains: PoT improves accuracy to 64.17\% from 63.42\% while reducing token consumption by 58.88\% (153K vs. 373K) on the \emph{Medium} subset. \textbf{When addressing complex numerical reasoning problems, PoT avoids precise numerical calculations by utilizing external tools (\ie{, Python interpreter}) and reduces the need for repetitive text-based reasoning, which is beneficial for most MLLMs.} 

However, we also observe that for certain reasoning-enhanced models (\ie{, Claude 3.7 Sonnet with 64K extended thinking and OpenAI o1}), due to the enforced requirement for long thought, they still engage in extensive text-based reasoning before generating code even in PoT setting, resulting in exceptionally high token consumption on the \emph{Hard} subset (4.48M and 2.12M), which is more than 10 times that of other MLLMs. To further investigate this, we added an IO baseline without any external prompts for reasoning-enhanced models on the \emph{Hard} subset. The IO group achieved the highest accuracy, which we attribute to the comprehensive built-in system prompts embedded in the tested proprietary models. \textbf{This highlights the need for future research to balance reasoning performance with the control of inefficient and redundant token generation, aiming to achieve a good trade-off between performance and cost, as well as to investigate whether PoT prompting can yield significant performance gains on open-source reasoning-enhanced models.}


\subsection{Error Analysis (RQ2)}
To better analyze the capabilities and limitations of MLLMs on \ours, we conduct a detailed error analysis for the Claude 3.7 Sonnet with 64K extended thinking in PoT setting. Error analysis is based on 100 sampled failure cases, which we categorize into three main error types, some of which involve compound errors. More details of error cases are provided in the Appendix.

\begin{itemize}[leftmargin=*]
\itemsep 0em
\item \textbf{Visual Perception Error} (30/100): The model incorrectly perceives, identifies, or interprets visual information from images, or mistakenly recognizes incorrect data, subsequently causing errors in calculations, broken reasoning chains, or incorrect conclusions.

\item \textbf{Knowledge Reasoning Error} (38/100): Due to insufficient domain-specific knowledge, the model exhibits logical confusion or conceptual misunderstandings during reasoning, leading to incorrect answers.

\item \textbf{Numerical Computation Error} (32/100): In problems involving mathematical operations and numerical reasoning, the model produces significant deviations from the correct answers due to errors in the calculation steps, precision control, or numerical hallucination.

\end{itemize}

\subsection{Visual Filtering for Reasoning (RQ3)}
As shown in \cref{tab:new_style}, when processing multi-image inputs containing distractor images, Qwen2.5-VL-72B demonstrates significantly lower accuracy across all difficulty levels compared to ground images scenarios. This finding aligns with conclusions from previous work~\cite{liu-etal-2024-mibench,sharma-etal-2024-losing}, indicating that irrelevant visual information substantially interferes with MLLMs' reasoning capabilities. In particular, the \emph{Medium} \emph{validation} set exhibits the most pronounced performance drop (77.78\% ground images vs. 64.73\% distractor images), attributed to two key characteristics: (1) moderate complexity making visual perception quality the performance bottleneck; (2) semantic relevance between distractors and questions increasing visual filtering difficulty.
To address distractor image interference, we propose a two-stage multimodal reasoning pipeline:
\begin{itemize} [leftmargin=*]
    \itemsep 0em 
    \item \textbf{Visual Filtering}: We first instruct MLLM to analyze the set of images $\mathcal{I}$ and the question $q$, assessing the relevance of the image (relevant/irrelevant). Irrelevant images are excluded from subsequent reasoning.
    \item \textbf{Enhanced Reasoning}: Then, the filtered subset $\mathcal{I}'$ and the question $q$ are input into the MLLM for the final reasoning. The system automatically reverts to the original inputs $\mathcal{I}$ if all images are mistakenly filtered.
\end{itemize}
\noindent \textbf{Does the pipeline help?} As illustrated in \cref{tab:new_style}, we evaluate Qwen2.5-VL-72B on the 207 English questions with distractor images of the \emph{Medium} validation set. Our method achieves an overall accuracy of 71.56\%, representing a 6.83 percentage points improvement over direct reasoning. This result is only 6.22 percentage points away from the ideal accuracy of the ground images scenarios (77.78\%). Detailed analysis reveals 73.43\% (152/207) ground image recognition accuracy during filtering. When correctly identified, the accuracy of these problems increases to 81.58\% (124/152). \textbf{This finding underscores the necessity to enhance the ability of MLLMs to filter out irrelevant image information, thereby strengthening their robustness in reasoning within more complex real-world scenarios.}

\subsection{Knowledge Augmentation (RQ4)}

To enhance the understanding and application capabilities of financial knowledge of MLLMs, we explore a method of refined knowledge augmentation to improve the performance of MLLM in domain-specific reasoning tasks.

\begin{itemize} [leftmargin=*]
\itemsep 0em
\item \textbf{Function Library Construction}: We annotate a comprehensive financial function library containing 3,133 Python functions from financial encyclopedia. Each function includes precise functional descriptions, parameter explanations, and step-by-step implementation code.

\item \textbf{MLLM-Instructed Knowledge Retrieval}: In financial problems with hybrid contexts, using short questions or full contexts for retrieval often fails to retrieve directly relevant knowledge~\cite{chen-etal-2023-beyond,peng2023check}. We observed that powerful MLLMs (\eg{, GPT-4o}) can effectively summarize rich semantic information from contexts. Therefore, we first prompt the MLLM to generate precise retrieval queries based on the context~\cite{li2025searcho1,verma2025planrag}. Then we use Contriever~\cite{izacardunsupervised} to retrieve relevant financial Python functions, based on the semantic similarity between the refined queries and functional descriptions.
    
\item \textbf{MLLM as Retrieval Judge}: Recent studies have shown that models are capable of judging the relevance of candidates retrieved for the question~\cite{guan2024amor}. In this setting, we first retrieved the Top-30 financial functions and then prompted the MLLM to select the Top-3 functions most useful to answer the question, if any.
    
\end{itemize}

\input{tables/table_rag_diff}

\noindent \textbf{Does knowledge augmentation help?} As shown in \cref{tab:rag_diff}, all evaluated MLLMs enhanced with financial function knowledge achieved significant improvements, ranging from 2.76 to 4.31 percentage points. Leveraging the improved retrieval efficiency enabled by \emph{MLLM-Instructed Knowledge Retrieval} and \emph{MLLM as Retrieval Judge}, the knowledge augmentation approach achieves greater performance, boosting the accuracy of Claude 3.7 Sonnet with 64K thinking to 86.29\%. Notably, Gemini 2.0 Flash Thinking, which has relatively weaker reasoning capabilities, also improved from 78.71\% to 83.02\%, approaching the performance of Claude 3.7 Sonnet (83.53\%) without knowledge augmentation. \textbf{The results further illustrate that refined domain-specific reasoning knowledge can significantly enhance the performance of MLLMs in complex reasoning tasks within expert domains.}


\subsection{Visual Parser with Reasoner (RQ5)}

In complex multimodal numerical reasoning tasks, single models often struggle to simultaneously achieve visual perception, knowledge reasoning, and numerical computation. To investigate the potential of model collaboration, we instruct the MLLM to act as the \textbf{Visual Parser}, responsible for carefully converting images into structured textual data. Then, an LRM acts as the \textbf{Reasoner}, performing multi-step numerical reasoning based on the textual context.

Specifically, we filter 1,160 tabular QA instances from FinMMR and utilize \textbf{GPT-4o} as the Visual Parser, instructing it to separate table headers or cells with vertical bars (\textbf{\textbar}) and rows with newlines. For the Reasoner component, in addition to \textbf{GPT-4o}, we evaluate several LRMs (\ie{, \textbf{Claude 3.7 Sonnet}, \textbf{Gemini 2.0 Flash Thinking}, \textbf{DeepSeek-R1}, and \textbf{OpenAI o1}}).

\begin{figure}[!t]
    \centering
    \includegraphics[width=\linewidth]{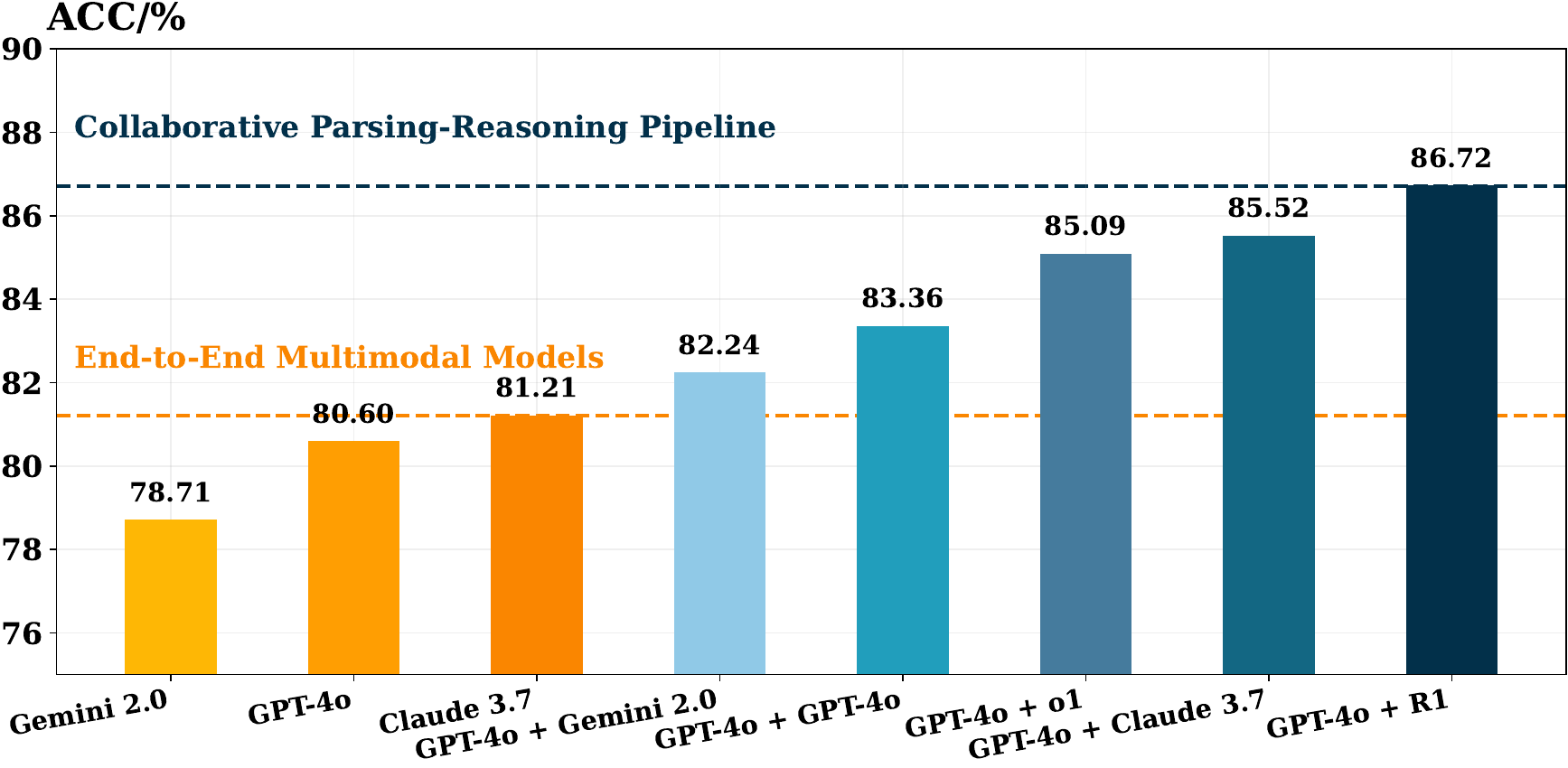}
    \caption{Results of model combinations and
individual models.}
    \label{fig:model_collaboration}
\end{figure}

\noindent \textbf{Does model collaboration help?} As shown in \cref{fig:model_collaboration}, the structured data from GPT-4o's visual parsing significantly enhances downstream reasoning. The individual model (\ie{, GPT-4o with PoT}) achieves an accuracy of 80.60\%, while the combination of models improves the accuracy to 86.72\% (\ie{, DeepSeek-R1 serving as the Reasoner with PoT}). Performance variance emerges across reasoning-enhanced models using identical visual inputs. Claude 3.7 Sonnet reaches 85.52\%, outperforming Gemini 2.0 Flash Thinking (82.24\%), confirming the decisive impact of text-based reasoning capabilities. \textbf{This evidences that model collaboration effectively compensates for individual model limitations through complementary strengths.}

%% file: tables/table_main_result.tex
\begin{table*}[!ht]

\centering
\fontsize{8pt}{10pt}\selectfont
\resizebox{\textwidth}{!}{%
\renewcommand{\arraystretch}{1.0}
\addtolength{\tabcolsep}{-0.1em}
\begin{tabular}{lrc rrrrrrrrrrrrrrr}
\toprule
\multirow{2}{*}{\textbf{Model}} & \multirow{2}{*}{\textbf{Size}}  & \multirow{2}{*}{\makecell[c]{\raisebox{-0.2em}{\textbf{Extended}}\\ \raisebox{-0.2em}{\textbf{thinking}}}} & \multicolumn{3}{c}{\textbf{Hard}} & ~ & \multicolumn{2}{c}{\textbf{Medium}} & ~ & \multicolumn{2}{c}{\textbf{Easy}} & ~ & \multicolumn{2}{c}{\textbf{Avg.}} & ~ & \multicolumn{2}{c}{\textbf{Token (M)}}\\
\cmidrule(lr){4-6} \cmidrule(lr){8-9} \cmidrule(lr){11-12} \cmidrule(lr){14-15} \cmidrule(lr){17-18}
~ & ~ & ~ & IO & CoT & PoT & & CoT & PoT & & CoT & PoT & & CoT & PoT & & CoT & PoT\\

\midrule
\multicolumn{14}{l}{\emph{\textbf{Proprietary MLLMs}}} \\

Claude 3.7 Sonnet & ~ & \textcolor{red}{\textcolor{darkgreen}{\ding{52}}} (64K) & \cellcolor{orange!45}\textbf{53.00} & \cellcolor{orange!45}\textbf{51.00} & \cellcolor{orange!45}\textbf{51.40} & & 62.50 & \cellcolor{orange!30}62.17 & & \cellcolor{orange!45} \textbf{78.50} & \cellcolor{orange!45}\textbf{78.50} & & \cellcolor{orange!45}\textbf{64.00} & \cellcolor{orange!45}\textbf{64.02} & & \cellcolor{blue!45}\textbf{8.51} & \cellcolor{blue!45}\textbf{11.25} \\

Claude 3.7 Sonnet & ~ & \textcolor{red!80}{\ding{56}} & \cellcolor{orange!30}49.80 & \cellcolor{orange!30}50.80 & \cellcolor{orange!30}48.50 & & 62.25 & 58.83 & & 77.00 & \cellcolor{orange!15}76.92 & & \cellcolor{orange!30}63.35 & 61.42 & & 0.99 & 0.89 \\

OpenAI o1 & ~ & \textcolor{darkgreen}{\ding{52}} & \cellcolor{orange!15}48.00 & 48.40 & 44.70 & & -- & -- & & -- & -- & & -- & -- & & \cellcolor{blue!15}2.52 & \cellcolor{blue!15}2.12 \\

GPT-4o & ~ & \textcolor{red!80}{\ding{56}} & -- & 45.40 & \cellcolor{orange!15}47.80 & & \cellcolor{orange!30}63.33 & 59.92 & & \cellcolor{orange!30}78.00 & 76.00 & & 62.24 & 61.24 & & 0.85 & 0.41 \\

Gemini 2.0 Pro & ~ & \textcolor{red!80}{\ding{56}} & -- & 46.50 & 47.30 & & 60.58 & 57.92 & & 75.50 & 75.67 & & 60.86 & 60.30 & & 0.85 & 0.45 \\

Gemini 2.0 Flash Thinking & ~ & \textcolor{darkgreen}{\ding{52}} & -- & 46.00 & 46.00 & & 60.75 & 56.58 & & 77.17 & 74.17 & & 61.31 & 58.92 & & 1.30 & 0.48 \\

Gemini 2.0 Flash & ~ & \textcolor{red!80}{\ding{56}} & -- & 44.40 & 45.90 & & 57.83 & 53.42 & & 74.92 & 73.75 & & 59.05 & 57.69 & & 0.79 & 0.43 \\

Grok 2 Vision & ~ & \textcolor{red!80}{\ding{56}} & -- & 27.80 & 25.50 & & 41.50 & 35.83 & & 73.08 & 72.83 & & 47.46 & 44.72 & & 1.13 & 0.60 \\

Qwen-Omni-Turbo & ~ & \textcolor{red!80}{\ding{56}} & -- & 17.50 & 27.30 & & 35.83 & 48.00 & & 57.50 & 61.67 & & 36.94 & 45.66 & & 0.90 & 0.42 \\

\midrule
\multicolumn{14}{l}{\emph{\textbf{Open-source MLLMs}}} \\

Llama 4 Maverick & 17B & \textcolor{red!80}{\ding{56}} & -- & \cellcolor{orange!15}48.70 & \cellcolor{orange!15}47.80 & & \cellcolor{orange!15}63.25 & 59.17 & & \cellcolor{orange!15}77.83 & \cellcolor{orange!30}77.83 & & \cellcolor{orange!15}63.26 & \cellcolor{orange!15}61.60 & & 0.88 & 0.47 \\

Qwen2.5-VL-72B & 72B & \textcolor{red!80}{\ding{56}} & -- & 43.30 & 46.20 & & \cellcolor{orange!45}\textbf{63.42} & \cellcolor{orange!45}\textbf{64.17} & & 77.42 & 75.83 & & 61.38 & \cellcolor{orange!30}62.07 & & 1.05 & 0.44 \\

InternVL2.5-78B & 78B & \textcolor{red!80}{\ding{56}} & -- & 37.40 & 44.00 & & 60.50 & \cellcolor{orange!15}61.17 & & 70.92 & 70.58 & & 56.27 & 58.58 & & -- & -- \\

QVQ-72B-Preview & 72B & \textcolor{darkgreen}{\ding{52}} & 43.30 & 40.30 & 6.20 & & 55.67 & 9.67 & & 75.42 & 12.42 & & 57.13 & 9.43 & & \cellcolor{blue!30}5.43 & \cellcolor{blue!30}5.70 \\

Pixtral Large & 124B & \textcolor{red!80}{\ding{56}} & -- & 19.70 & 25.00 & & 39.83 & 39.75 & & 70.00 & 70.17 & & 43.18 & 44.97 & & 1.15 & 0.75 \\

Gemma 3 27B & 27B & \textcolor{red!80}{\ding{56}} & -- & 23.40 & 22.30 & & 45.17 & 36.42 & & 69.08 & 61.58 & & 45.88 & 40.10 & & 0.97 & 0.47 \\

Mistral Small 3.1 & 24B & \textcolor{red!80}{\ding{56}} & -- & 19.70 & 15.20 & & 38.42 & 29.75 & & 67.67 & 49.42 & & 41.93 & 31.46 & & 1.15 & 0.60 \\








\bottomrule
\end{tabular}}
\caption{Results of different models using IO, CoT and PoT prompting methods on the \emph{test} set of FinMMR. We use the best Accuracy on the \emph{Hard} subset as the ranking indicator of model performance. The results underscore the superior performance of reasoning-enhanced MLLM (\ie{, Claude 3.7 Sonnet with 64K extended thinking}) with PoT in complex multimodal numerical reasoning task.}
\label{tab:main_result}
\end{table*}

%% file: tables/table_vision_filter.tex
\begin{table*}[htbp]
    \centering
    \resizebox{\textwidth}{!}{ 
    \begin{tabular}{lcccccc}
    \toprule
    & \multicolumn{2}{c}{\textbf{Test}} & & \multicolumn{2}{c}{\textbf{Validation}} & \\
    \cmidrule(lr){2-3}\cmidrule(lr){5-6}
    \textbf{Subset} & \textbf{Ground Images (\%)} & \textbf{Distractor Images (\%)} & \textbf{Degradation}
    & \textbf{Ground Images (\%)} & \textbf{Distractor Images (\%)} & \textbf{Degradation}\\
    \midrule
    \textbf{Hard}   & 57.18 & 47.23 & \textcolor{blue}{$\downarrow$ 9.95} & 56.74 & 45.58 & \textcolor{blue}{$\downarrow$ 11.16} \\
    \textbf{Medium} & 73.01 & 61.36 & \textcolor{blue}{$\downarrow$ 11.65}  & 77.78 & 64.73 & \textcolor{blue}{$\downarrow$ 12.35}  \\
    \textbf{Easy}   & 61.59 & 53.64 & \textcolor{blue}{$\downarrow$ 7.95}  & 60.61 & 51.52 & \textcolor{blue}{$\downarrow$ 9.09}  \\
    \midrule
    \multicolumn{7}{l}{\textbf{The improvement achieved by the filtering-reasoning pipeline on the \emph{Medium} \emph{validation} set}: 64.73 $\rightarrow$ 71.56 \quad \textcolor{red}{$\uparrow$ \textbf{6.83}}} \\
    \bottomrule
    \end{tabular}%
    }
    \caption{Degradation of Qwen2.5-VL-72B on all subsets due to distractor images and improvement achieved by the filtering-reasoning pipeline on the \emph{Medium} \emph{validation} set in PoT setting.}
    \label{tab:new_style}
\end{table*}

%% file: tables/table_rag_diff.tex
\begin{table}[!t]
    \centering
    \resizebox{\linewidth}{!}{
    \begin{tabular}{lll}
        \toprule
        Setting & PoT & RAG with PoT\\
        \midrule 
        
        \noalign{\vskip 0ex}
        
        \noalign{\vskip 0ex}
        \noalign{\vskip 0ex}
         Gemini 2.0 Flash Thinking & 78.71  & 83.02 \up{4.31}\\
         GPT-4o & 80.60  & 83.62 \up{3.02}\\
         Claude 3.7 Sonnet & 81.21  & 85.43 \up{4.22}\\
         Claude 3.7 Sonnet (64K) & 83.53  & 86.29 \up{2.76}\\
        
        \bottomrule
    \end{tabular}
    }
    \caption{Improvements of different MLLMs with knowledge augmentation on the 1,160 instances of \ours in PoT setting.}
    \label{tab:rag_diff}
\end{table}

%% file: sec/6_conclusion.tex
\section{Conclusion}
We introduce \ours, a multimodal, comprehensive, and challenging benchmark for evaluating the financial numerical reasoning capabilities of MLLMs. \ours challenges MLLMs' intricate visual perception, specialized knowledge reasoning, and accurate numerical computation through its rich images, comprehensive subdomains, and complex formulas embedded in each multimodal financial question. The evaluation results reveal that 15 proprietary and open-source MLLMs still struggle with complex multimodal reasoning tasks in specialized domains. \ours highlights the bottlenecks of MLLMs and the need for continuous improvements, including reasoning-enhanced training, tool use, refined structured knowledge augmentation and  model combinations, allowing models to perform expert-level reasoning tasks closer to the real-world scenarios.

%% file: sec/acknowledgements.tex
\section*{Acknowledgements}
This work is supported by the National Natural Science Foundation of China (Grant Nos. 62176026, 62473271), the Beijing Natural Science Foundation (Grant No. QY24214), the BUPT Innovation and Entrepreneurship Support Program (Grant Nos. 2025-YC-A033, 2025-YC-A042), and data support from Hithink RoyalFlush Information Network Co., Ltd. This work is also supported by the Engineering Research Center of Information Networks, Ministry of Education, China. We would also like to thank the anonymous reviewers and area chairs for constructive discussions and feedback.

%% file: error/main.tex
\section{Error Analysis}
We exhibit representative examples for the three primary error categories: \textbf{Visual Perception Error} (\cref{fig:err1}), 
\textbf{Knowledge Reasoning Error} with two cases (\cref{fig:err2}, \cref{fig:err3}), 
and \textbf{Numerical Computation Error} (\cref{fig:err4}). These errors were observed in the outputs of Claude 3.7 Sonnet with 64K extended thinking in PoT setting.

These error cases highlight critical limitations in current multimodal large language models (MLLMs). \textbf{Visual Perception} errors primarily stem from the misinterpretation of complex visual data (such as charts) or failures to accurately extract numerical information from tables. \textbf{Knowledge Reasoning} errors result from either incorrect factual knowledge stored in the model or flawed logical reasoning processes. \textbf{Numerical Computation} errors occur due to inaccuracies in the model's internal computation mechanisms during reasoning. Consequently, FinMMR underscores the need for improvements in three core capabilities: intricate visual perception, specialized knowledge reasoning, and accurate numerical computation.

%% file: augmentation/main.tex
\section{Augmentation Methods}

\subsection{Visual Filtering for Reasoning}
\cref{fig:filter1} illustrates a representative example demonstrating our proposed two-stage filtering-reasoning pipeline applied to a multimodal numerical reasoning task.
Specifically, the task requires calculating fifth-year sales revenue for the respiratory category, comparing it with third-year data, and computing the growth rate.
Among the three input images, Images 1 and 2 contain task-relevant tables, while Image 3 is an irrelevant distractor.
Without visual filtering, although the model correctly avoided extracting numerical information from the irrelevant image, it misinterpreted values from relevant tables, resulting in a significant deviation from the reference answer.
When implementing our pipeline, the MLLM first assesses image relevance, correctly identifying Images 1 and 2 as \texttt{[USEFUL]} and Image 3 as \texttt{[USELESS]}, thus eliminating distractors from downstream reasoning.

\subsection{Knowledge Augmentation}
\cref{fig:filter2} presents a representative financial multimodal reasoning task, demonstrating the efficacy of our domain knowledge augmentation method. Specifically, the task requires calculating the impact on net income from switching inventory accounting methods from LIFO to FIFO, based on a provided financial table.
In the baseline without augmentation (``Original Output''), the model incorrectly treated the entire 2014 LIFO reserve as income, violating U.S. GAAP which stipulates that \textbf{only changes in LIFO reserve} impact current-year net income. This caused significant income overstatement. Conversely, with our augmentation approach (``Augmented Output''), the model correctly computed the effect using the \textbf{year-over-year change in LIFO reserve} with proper tax effect adjustment, achieving accurate income quantification.
This case highlights the critical importance of domain knowledge augmentation for MLLMs. Complementing this, \cref{fig:Financial Function} illustrates a representative financial function retrieved from our library.


\subsection{Visual Parser with Reasoner}
\cref{fig:filter3} demonstrates our two-stage parsing-reasoning pipeline comprising a \textbf{Visual Parser} and a \textbf{Reasoner}. The task requires calculating anticipated portfolio returns under two economic scenarios using probability-weighted returns from a tabular image.
We first instruct GPT-4o (as Visual Parser) to convert tabular data into structured markdown format. This structured output then enables a large reasoning model (LRM) to perform numerical reasoning. Compared to directly reasoning on raw visual inputs, GPT-4o's parsed data significantly enhances the downstream LRM's accuracy, underscoring the efficacy of model collaboration in multimodal reasoning.


%% file: prompt_experiment/main.tex
\section{Prompt Templates}
\subsection{Chain-of-Thought (CoT)}
The CoT prompt directs MLLMs to decompose complex reasoning tasks into sequential intermediate steps. This approach enhances performance on mathematical, logical, and analytical problems 
by requiring explicit articulation of each reasoning stage before final answer generation. See \cref{prompt:CoT} for the complete template.

\subsection{Program-of-Thought (PoT)}
The PoT prompt guides MLLMs to formulate solutions as executable code. By translating problems into programs with variables, functions, and logical operations, 
this approach enables precise computation and transparent reasoning, particularly effective for quantitative tasks. The complete template is provided in \cref{prompt:PoT}.

\subsection{Retrieval-Augmented Generation (RAG)}
The RAG prompt integrates retrieval of external knowledge with generative reasoning. This hybrid approach enables MLLMs to supplement parametric knowledge with dynamically retrieved information, enhancing answer accuracy for knowledge-intensive tasks. See \cref{prompt:RAG} for implementation details.

\subsection{Function Relevance Judgment}
This prompt evaluates the utility of retrieved financial functions for answering specific questions. It assesses both relevance and reliability, permitting only qualified knowledge to inform downstream reasoning. The complete template is available in \cref{prompt:judge}.

\subsection{Visual Relevance Judgment}
This prompt assesses image relevance for financial problem-solving. It classifies each input image as \texttt{[USEFUL]} or \texttt{[USELESS]}, with uncertainty favoring retention to prevent critical information loss. See \cref{prompt:filter} for details.

\section{Experimental Details}

\subsection{Model Specifications}
\cref{tab:ai-models} details the 15 MLLMs evaluated on \ours, covering models from 8 leading organizations. Each entry includes the official model identifier to ensure full experimental reproducibility.

\subsection{Validation Set Results}
\cref{tab:main_result} presents the accuracy metrics of 12 MLLMs on the \textbf{validation} set of \ours across three difficulty tiers. Models are evaluated using both Chain-of-Thought (CoT) and Program-of-Thought (PoT) prompting methods under standardized conditions.

%% file: error/error_type.tex
\begin{figure}[!b]
    \centering
    \includegraphics[width=0.8\textwidth]{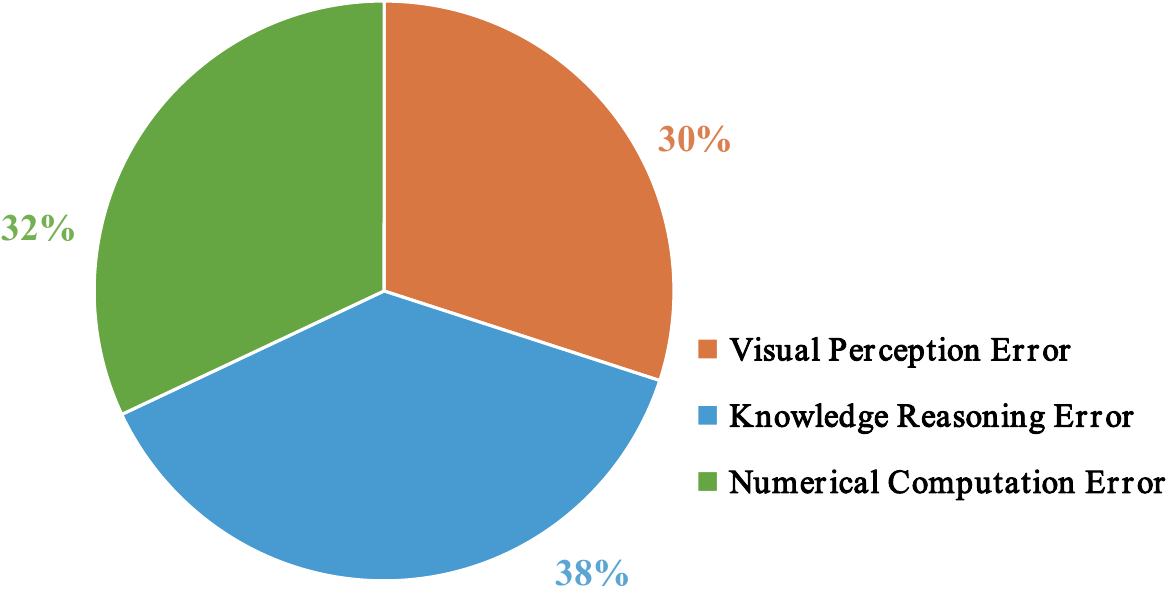}
    \caption{Error Distribution: Claude 3.7 Sonnet on FinMMR}
    \label{fig:error_type}
\end{figure}

%% file: error/visual_perception_error.tex
\begin{figure}[!t]
    \centering
    \includegraphics[width = 0.8\linewidth]{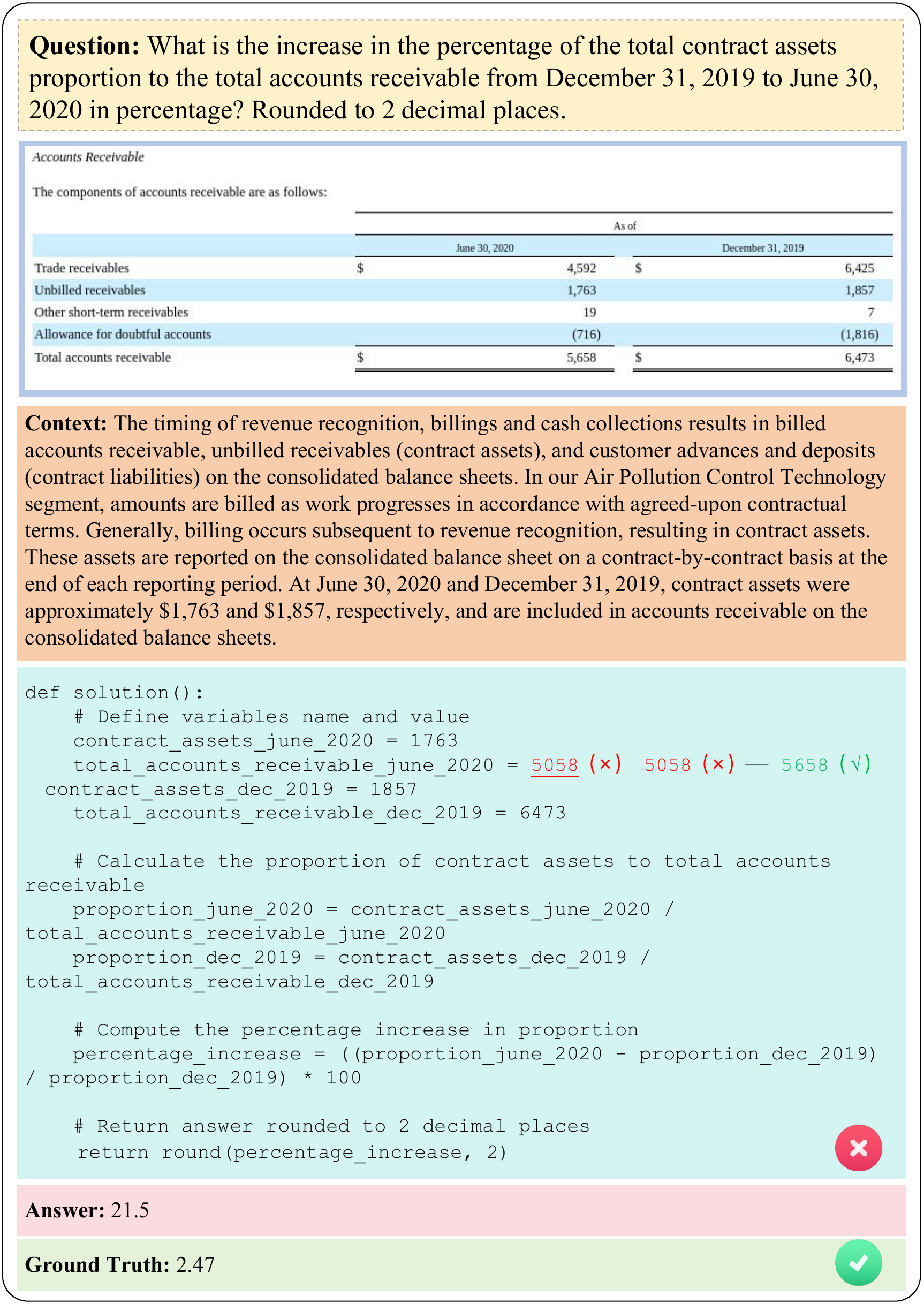}
    \caption{Visual Perception Error Case}
    \label{fig:err1}
\end{figure}

%% file: error/knowledge_reasoning_error1.tex
\begin{figure}[!t]
    \centering
    \includegraphics[width = 0.8\linewidth]{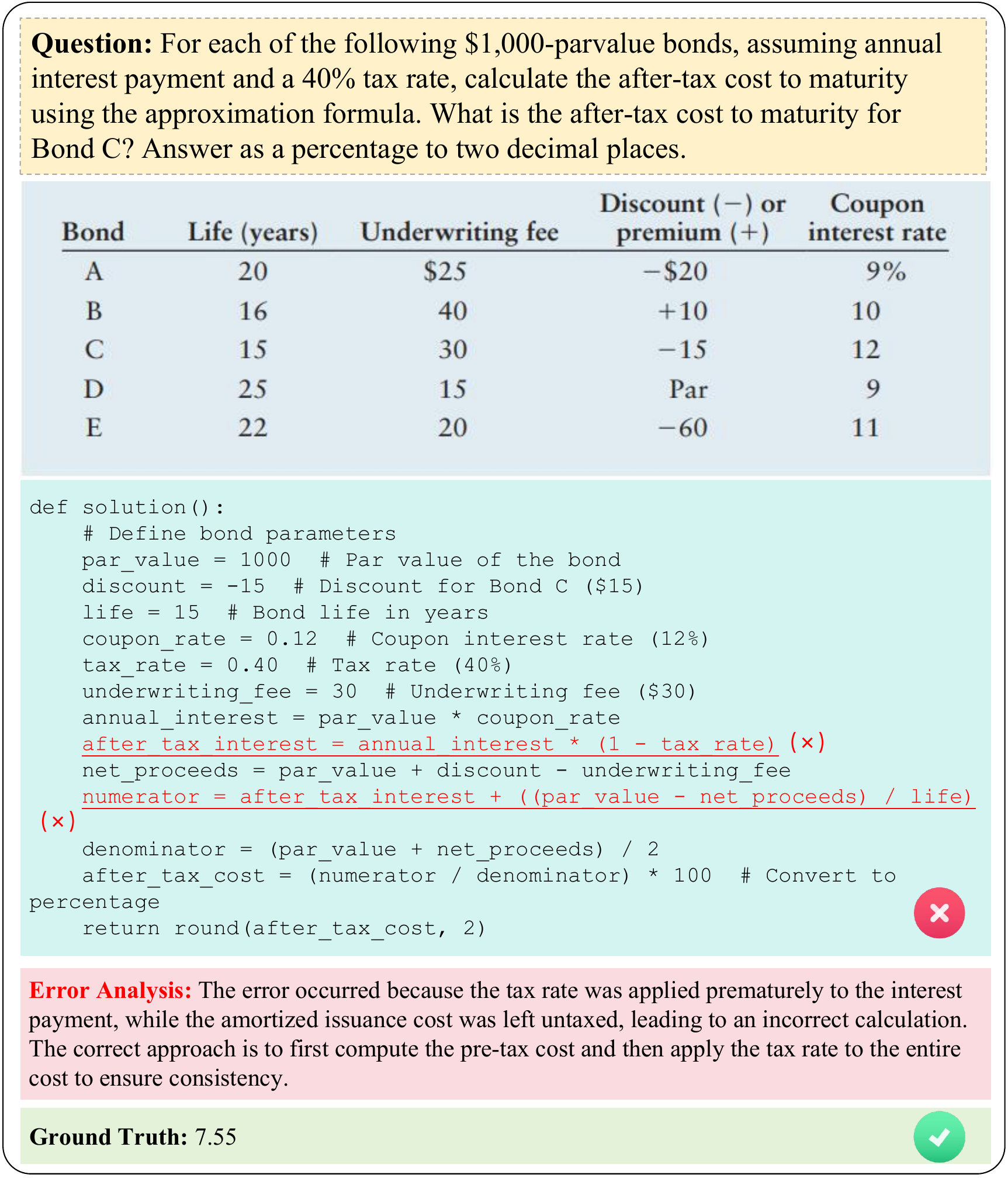}
    \caption{Knowledge Reasoning Error Case 1}
    \label{fig:err2}
\end{figure}

%% file: error/knowledge_reasoning_error2.tex
\begin{figure}[!t]
    \centering
    \includegraphics[width = 0.8\linewidth]{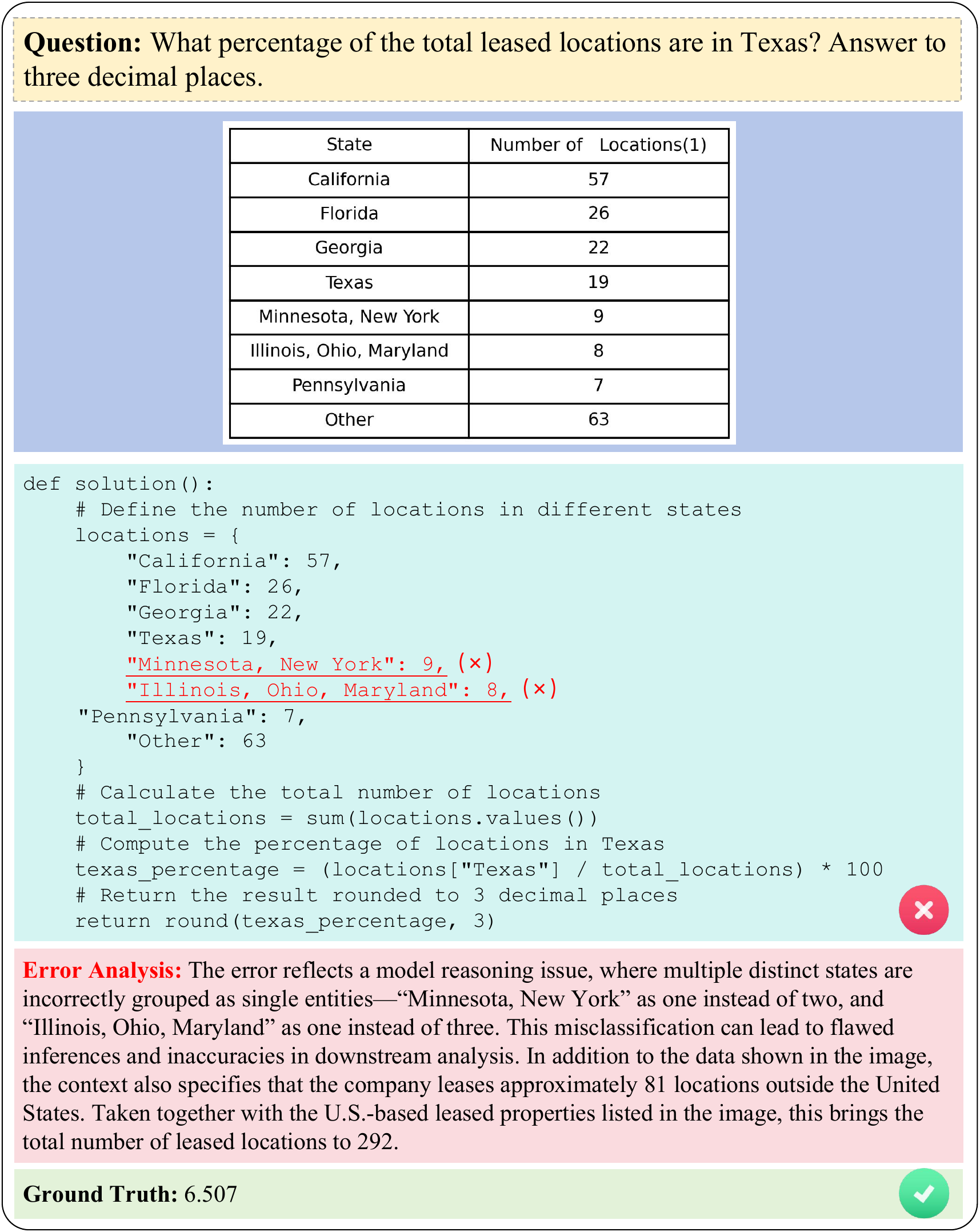}
    \caption{Knowledge Reasoning Error Case 2}
    \label{fig:err3}
\end{figure}

%% file: error/numerical_calculation_errors.tex
\begin{figure}[!t]
    \centering
    \includegraphics[width = 0.8\linewidth]{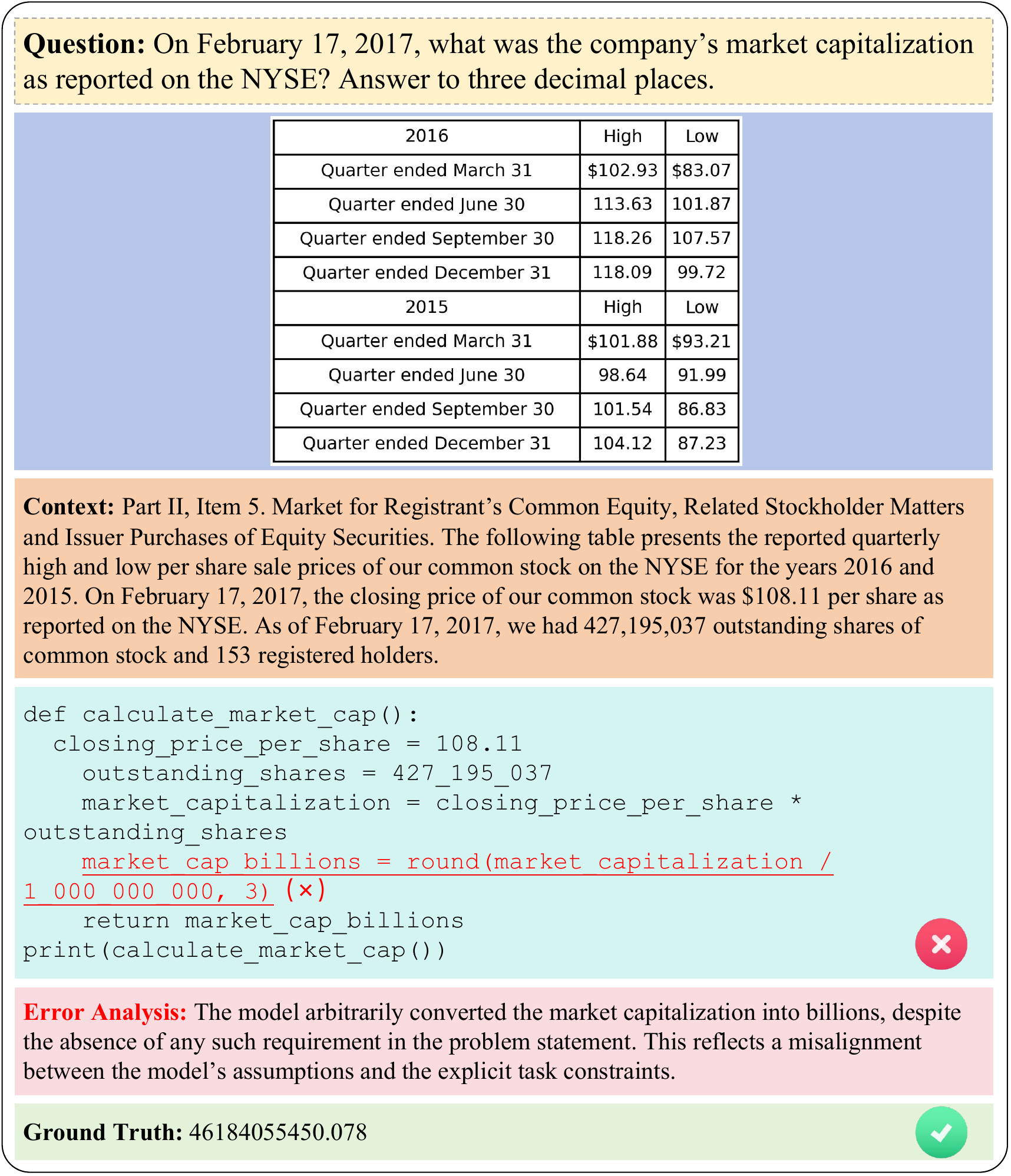}
    \caption{Numerical Computation Error Case}
    \label{fig:err4}
\end{figure} 

%% file: augmentation/pic_insert.tex
\onecolumn
\begin{figure}
    \centering
    \includegraphics[width=0.65\linewidth]{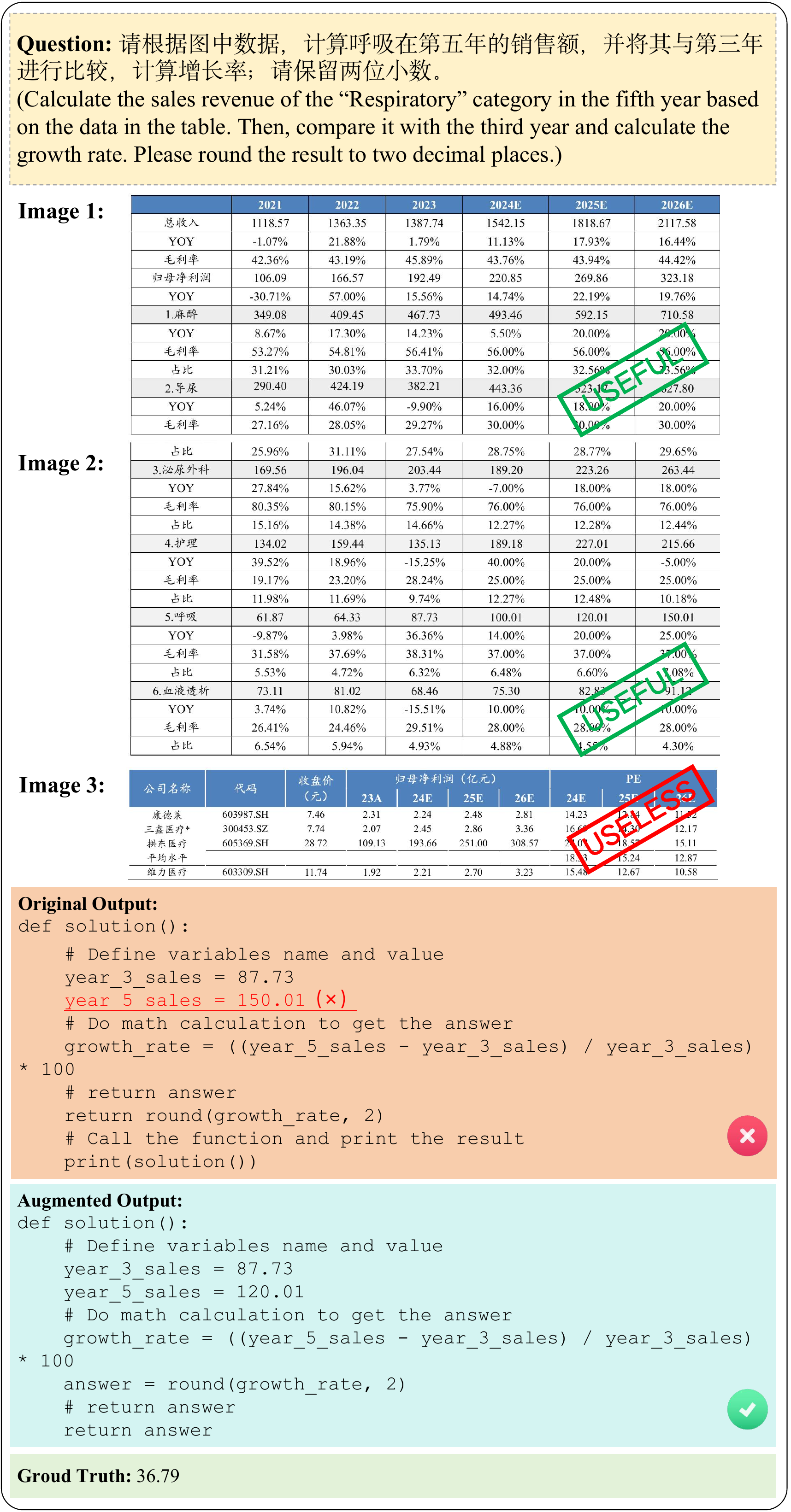}
    \caption{Example of Filtering-Reasoning Pipeline}
    \label{fig:filter1}
\end{figure}

\onecolumn
\begin{figure}
    \centering
    \includegraphics[width=0.65\linewidth]{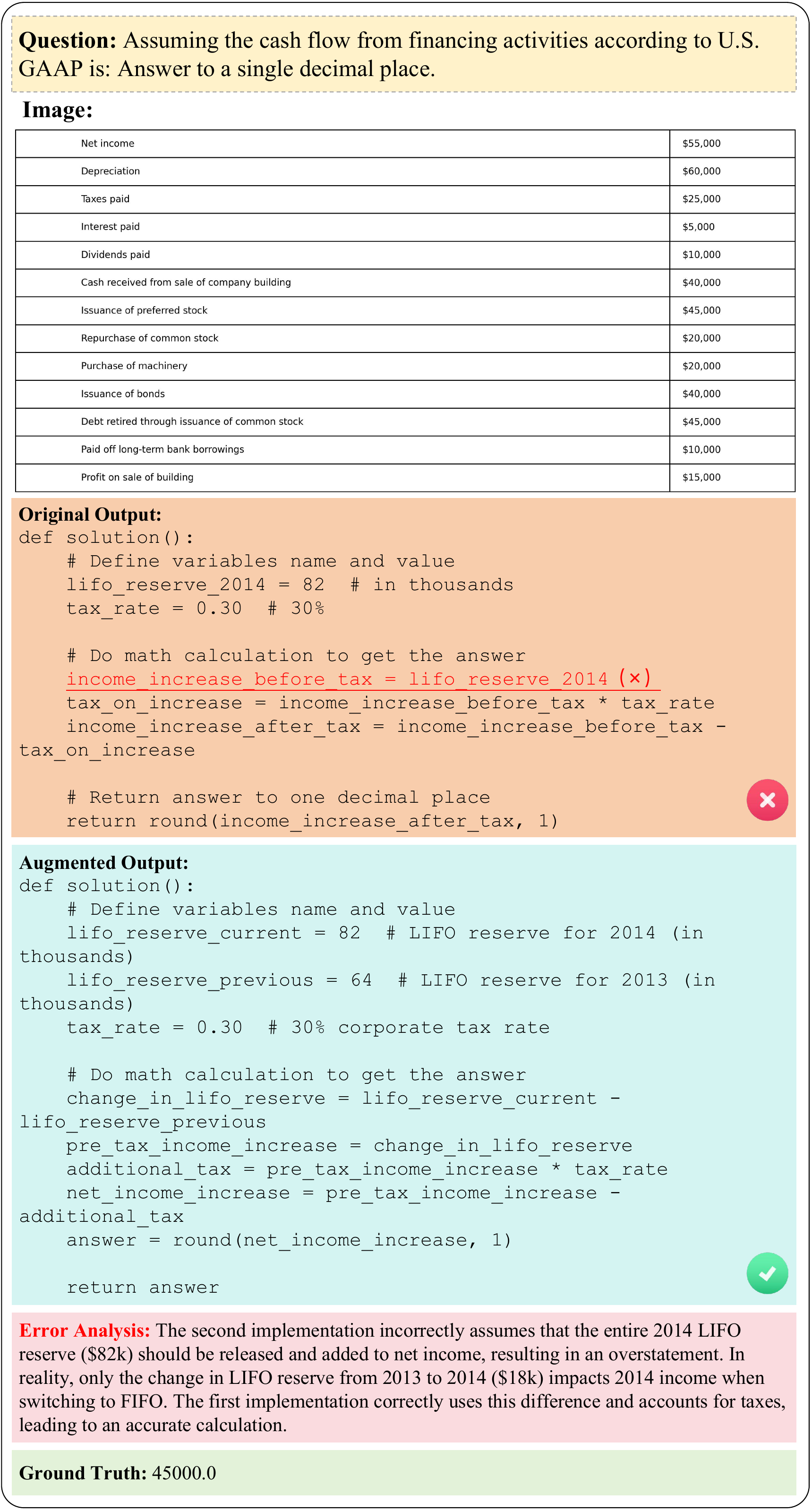}
    \caption{Example of Knowledge Augmentation}
    \label{fig:filter2}
\end{figure}

\onecolumn
\begin{figure}
    \centering
    \includegraphics[width=0.8\linewidth]{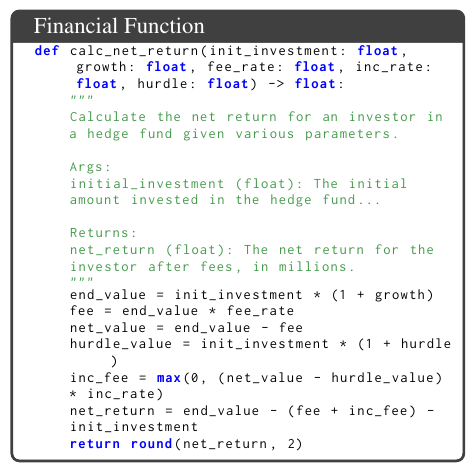}
    \caption{Example of Retrieved Financial Function}
    \label{fig:Financial Function}
\end{figure}

\onecolumn
\begin{figure}[!t]
    \centering
    \includegraphics[width=0.8\linewidth]{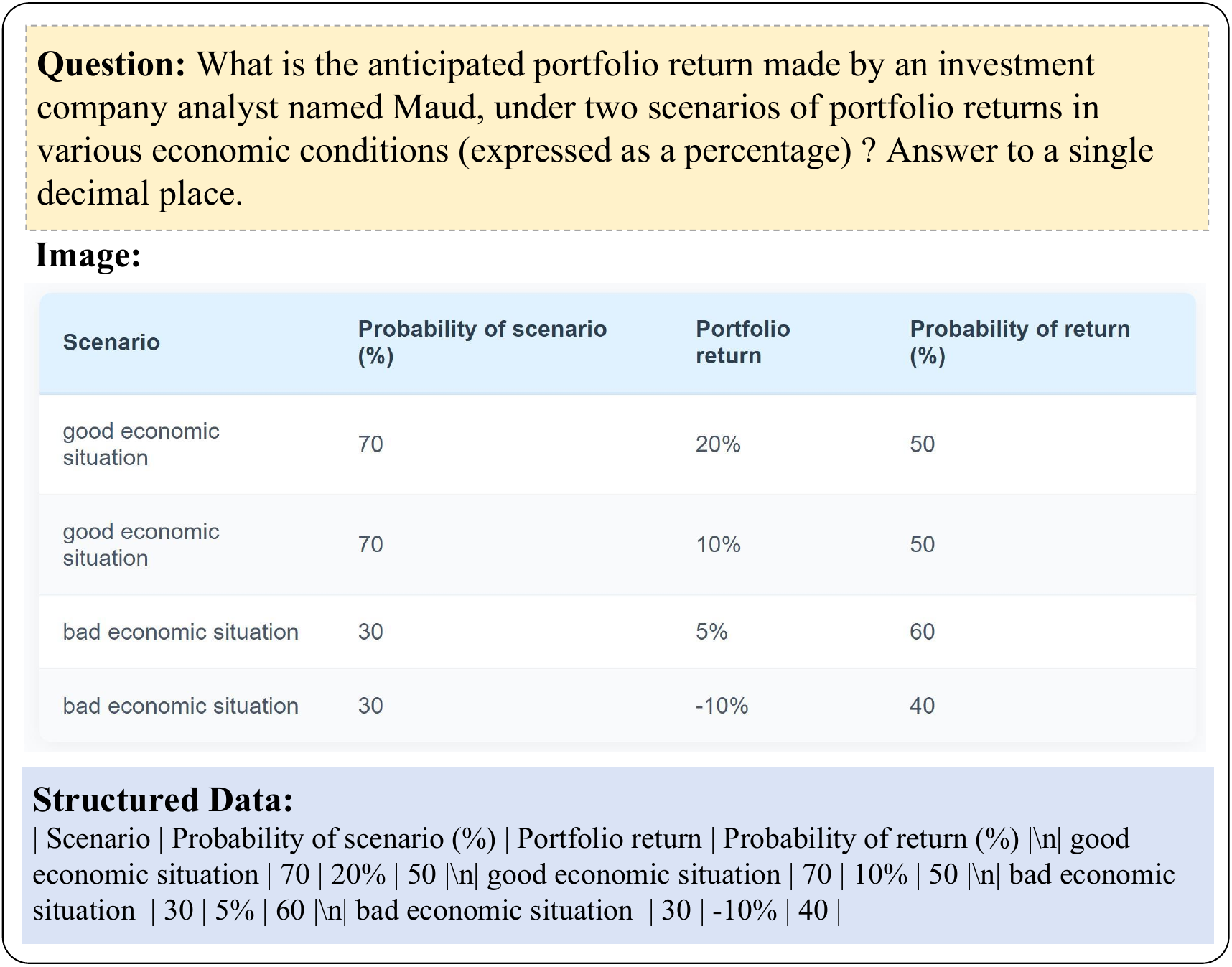}
    \caption{Example of Parsing-Reasoning Pipeline}
    \label{fig:filter3}
\end{figure}

%% file: prompt_experiment/table_model.tex
\onecolumn
\begin{table}
\centering
\begin{tabular}{lll}
\toprule
\textbf{Model} & \textbf{Organization} & \textbf{Source} \\
\midrule
Claude 3.7 Sonnet (Thinking) & Anthropic & claude-3-7-sonnet-20250219 \\
Claude 3.7 Sonnet & Anthropic & claude-3-7-sonnet-20250219 \\
\hdashline\noalign{\vskip 0.5ex}
GPT-4o & OpenAI & gpt-4o-2024-11-20 \\
OpenAI o1 & OpenAI & o1-2024-12-17 \\
\hdashline\noalign{\vskip 0.5ex}
Gemini 2.0 Flash Thinking & Google DeepMind & gemini-2.0-flash-thinking-exp-01-21 \\
Gemini 2.0 Pro & Google DeepMind & gemini-2.0-pro-exp-02-05 \\
Gemini 2.0 Flash & Google DeepMind & gemini-2.0-flash \\
Gemma 3 27B & Google DeepMind & gemma-3-27b-it \\
\hdashline\noalign{\vskip 0.5ex}
Qwen-Omni-Turbo & Alibaba Qwen & qwen-omni-turbo-2025-01-19 \\
Qwen2.5-VL-72B & Alibaba Qwen & qwen2.5-vl-72b-instruct \\
QvQ-72B-Preview & Alibaba Qwen & qvq-72b-preview \\
\hdashline\noalign{\vskip 0.5ex}
Pixtral Large & MistralAI & pixtral-large-latest \\
Mistral Small 3.1 & Mistral AI & mistral-small-3.1-24b-instruct \\
\hdashline\noalign{\vskip 0.5ex}
InternVL2.5-78B & OpenGVLab & OpenGVLab/InternVL2\_5-78B \\
Grok 2 Vision & xAI & grok-2-vision-1212 \\
Llama 4 Maverick & AI@Meta & llama-4-maverick \\

\bottomrule
\end{tabular}
\caption{Specifications of MLLMs Evaluated on \ours.}
\label{tab:ai-models}
\end{table}

%% file: prompt_experiment/table_validation_result.tex
\begin{table*}[!ht]

\centering
\fontsize{8pt}{10pt}\selectfont
\resizebox{\textwidth}{!}{%
\renewcommand{\arraystretch}{1.0}
\addtolength{\tabcolsep}{-0.1em}
\begin{tabular}{lrc rrrrrrrrrrrrrr}
\toprule
\multirow{2}{*}{\textbf{Model}} & \multirow{2}{*}{\textbf{Size}}  & \multirow{2}{*}{\textbf{\makecell[c]{\raisebox{-0.2em}{Extended}\\ \raisebox{-0.2em}{thinking}}}} & \multicolumn{2}{c}{\textbf{Hard}} & ~ & \multicolumn{2}{c}{\textbf{Medium}} & ~ & \multicolumn{2}{c}{\textbf{Easy}} & ~ & \multicolumn{2}{c}{\textbf{Avg.}} & ~ & \multicolumn{2}{c}{\textbf{Token (M)}}\\
\cmidrule(lr){4-5} \cmidrule(lr){7-8} \cmidrule(lr){10-11} \cmidrule(lr){13-14} \cmidrule(lr){16-17}
~ & ~ & ~ & CoT & PoT & & CoT & PoT & & CoT & PoT & & CoT & PoT & & CoT & PoT\\
\midrule

Claude 3.7 Sonnet & ~ & \textcolor{red}{\textcolor{darkgreen}{\ding{52}}} (64K) & \cellcolor{orange!45}\textbf{50.67} & \cellcolor{orange!45}\textbf{50.33} & & \cellcolor{orange!30}66.00 & \cellcolor{orange!15}63.00 & & \cellcolor{orange!45}\textbf{75.67} & \cellcolor{orange!45}\textbf{75.33} & & \cellcolor{orange!45}\textbf{64.11} & \cellcolor{orange!45}\textbf{62.89} & & \cellcolor{blue!45}\textbf{2.44} & \cellcolor{blue!45}\textbf{3.02} \\

Claude 3.7 Sonnet & ~ & \textcolor{red!80}{\ding{56}} &\cellcolor{orange!30} 50.33 & \cellcolor{orange!30}49.00 & & \cellcolor{orange!15}63.67 & 60.00 & & \cellcolor{orange!30}75.33 &\cellcolor{orange!30} 74.00 & & \cellcolor{orange!30}63.11 & 61.00 & & 0.27 & 0.24 \\

Qwen2.5-VL-72B & 72B & \textcolor{red!80}{\ding{56}} & 41.33 &\cellcolor{orange!15} 48.67 & & \cellcolor{orange!30}66.00 & \cellcolor{orange!45}\textbf{64.00} & & 77.00 & 70.00 & & 61.44 & \cellcolor{orange!15}61.45 & & 0.28 & 0.12 \\

GPT-4o & ~ & \textcolor{red!80}{\ding{56}} & 46.00 & 47.67 & &\cellcolor{orange!30} 66.00 & \cellcolor{orange!30}63.67 & & 73.33 & \cellcolor{orange!15}73.67 & & 61.78 &\cellcolor{orange!30} 61.67 & & 0.23 & 0.11 \\ 

InternVL2.5-78B & 78B & \textcolor{red!80}{\ding{56}} & 38.00 & 47.67 & & 57.00 & 60.67 & & 69.00 & 69.33 & & 54.67 & 59.22 & & -- & -- \\

Gemini 2.0 Flash Thinking & ~ & \textcolor{darkgreen}{\ding{52}} &\cellcolor{orange!15} 46.67 & 46.67 & & \cellcolor{orange!45}\textbf{67.00} & 60.33 & & \cellcolor{orange!15}73.67 & 73.33 & & \cellcolor{orange!15}62.45 & 59.00 & & \cellcolor{blue!15}0.33 & 0.13 \\ 

Gemini 2.0 Flash & ~ & \textcolor{red!80}{\ding{56}} & \cellcolor{orange!15}46.67 & 46.00 & & 63.33 & 55.67 & & 70.00 & 71.33 & & 60.00 & 57.67 & & 0.31 & 0.12 \\

Gemini 2.0 Pro & ~ & \textcolor{red!80}{\ding{56}} & 45.00 & 44.00 & & 61.33 & 62.00 & & 71.33 & 71.33 & & 59.22 & 59.78 & & 0.23 & 0.12 \\ 

OpenAI o1 & ~ & \textcolor{darkgreen}{\ding{52}} & -- & 45.00 & & -- & -- & & -- & -- & & -- & -- & & -- & \cellcolor{blue!15}0.61 \\

QVQ-72B-Preview & 72B & \textcolor{darkgreen}{\ding{52}} & 40.33 & 6.00 & & 58.33 & 8.33 & & 73.33 & 11.00 & & 57.33 & 8.44 & &\cellcolor{blue!30} 1.46 & \cellcolor{blue!30}1.54 \\ 

Qwen-Omni-Turbo & ~ & \textcolor{red!80}{\ding{56}} & 18.33 & 30.33 & & 35.67 & 45.33 & & 53.67 & 60.67 & & 35.89 & 45.44 & & 0.24 & 0.11 \\

Grok 2 Vision & ~ & \textcolor{red!80}{\ding{56}} & 27.67 & 26.33 & & 42.67 & 33.67 & & 72.00 & 73.00 & & 47.45 & 44.33 & & 0.30 & 0.16 \\ 

Pixtral Large & 124B & \textcolor{red!80}{\ding{56}} & 25.00 & 27.00 & & 38.67 & 38.00 & & 65.33 & 67.33 & & 43.00 & 44.11 & & 0.28 & 0.20 \\ 

    \bottomrule
    \end{tabular}
    }
\caption{Results of different models using CoT and PoT prompting methods on the \emph{validation} set of FinMMR. We use the best Accuracy on the \emph{Hard} subset as the ranking indicator of model performance.}
\label{tab:validation_result}
\end{table*}

%% file: prompt_experiment/prompt_cot.tex
\begin{figure}[!t]
\centering
\includegraphics[width=\linewidth]{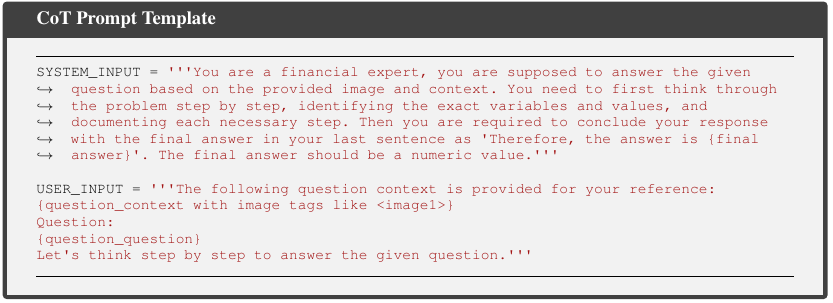}
\caption{CoT Prompt Template}
\label{prompt:CoT}
\end{figure}

%% file: prompt_experiment/prompt_pot.tex
\begin{figure}[!t]
\centering
\includegraphics[width=\linewidth]{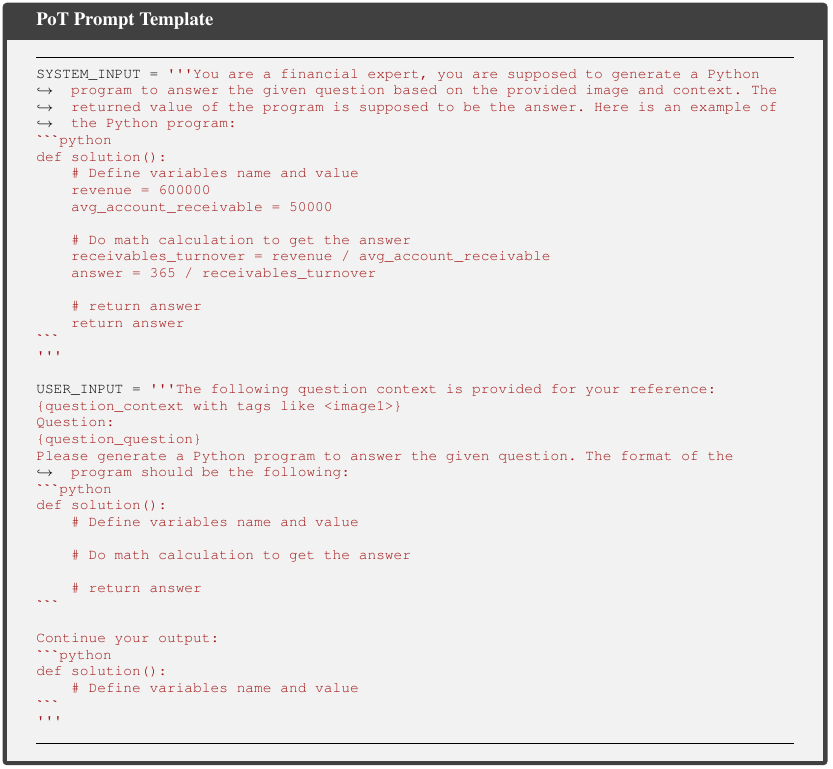}
\caption{CoT Prompt Template}
\label{prompt:PoT}
\end{figure}

%% file: prompt_experiment/prompt_rag.tex
\begin{figure}[!t]
\centering
\includegraphics[width=\linewidth]{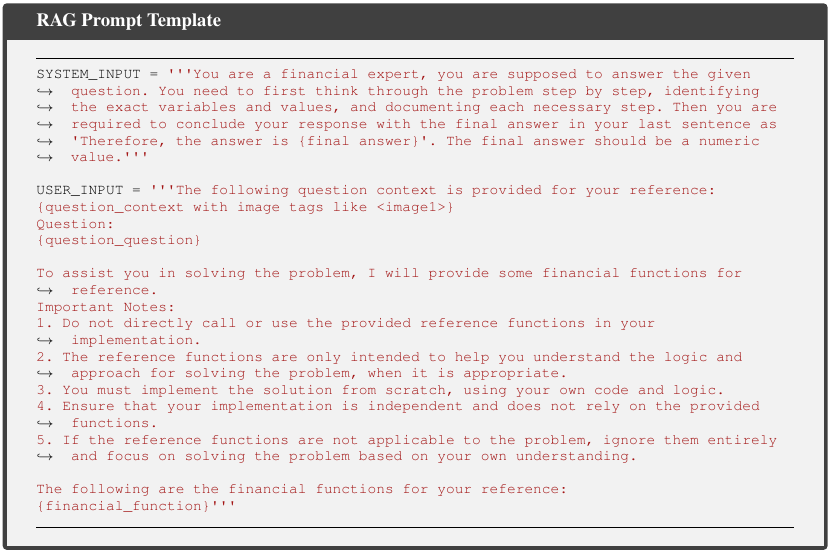}
\caption{CoT Prompt Template}
\label{prompt:RAG}
\end{figure}

%% file: prompt_experiment/prompt_judge_useful_function.tex
\begin{figure}[!t]
\centering
\includegraphics[width=\linewidth]{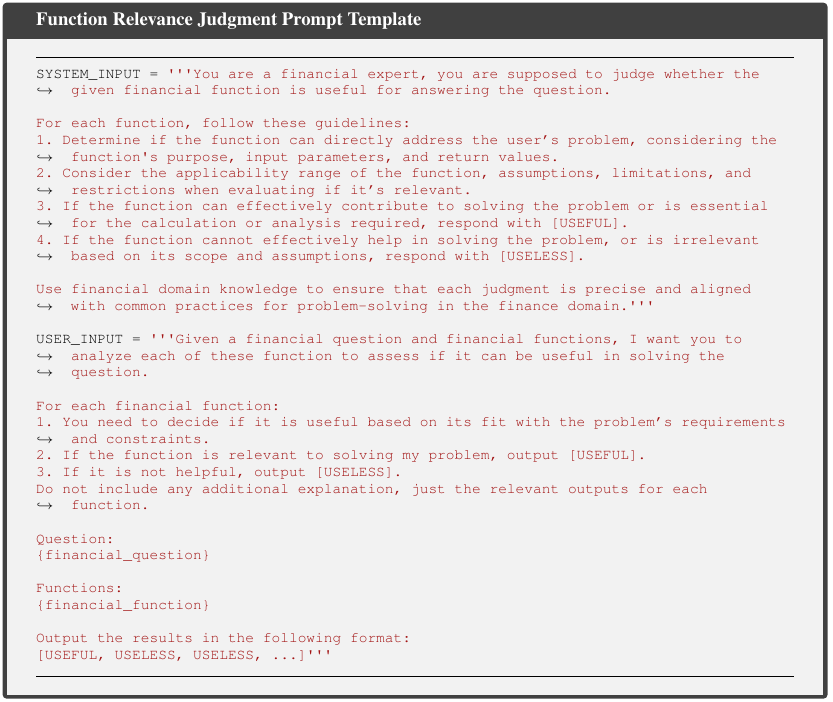}
\caption{CoT Prompt Template}
\label{prompt:judge}
\end{figure}

%% file: prompt_experiment/prompt_judge_correlation_of_image.tex
\begin{figure}[!t]
\centering
\includegraphics[width=\linewidth]{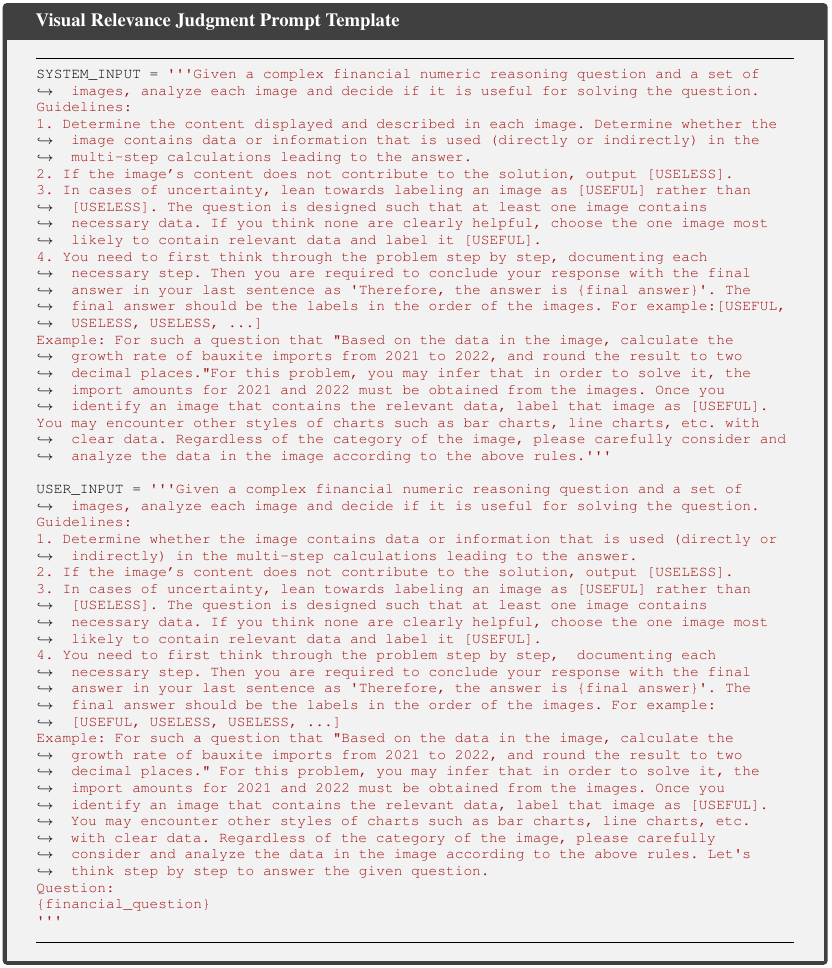}
\caption{CoT Prompt Template}
\label{prompt:filter}
\end{figure}